\documentclass[11pt]{article}
\usepackage{enumitem}

\usepackage[final]{acl}

\usepackage{times}
\usepackage{latexsym}
\usepackage{ulem} 

\usepackage[T1]{fontenc}

\usepackage[utf8]{inputenc}

\usepackage{microtype}

\usepackage{inconsolata}

\usepackage{graphicx}
\usepackage{comment}
\usepackage{hyperref}
\usepackage{url}
\usepackage{algorithm}
\usepackage{algpseudocode}
\usepackage{subcaption}
\usepackage{booktabs}
\usepackage{multirow}
\usepackage[most]{tcolorbox}
\tcbuselibrary{skins,breakable}
\usepackage{tabularx}
\usepackage{array}
\usepackage{xcolor}
\usepackage{enumitem}
\usepackage{capt-of}

\newtcolorbox{promptbox}[1][]{
    colback=black!5,      
    colframe=black!75,    
    fonttitle=\bfseries,  
    title=Prompt,         
    arc=2mm,              
    boxrule=1pt,          
    width=\textwidth,    
    #1                    
}

\newtcolorbox{problembox}{
  enhanced,
  colback=gray!10,
  colframe=gray!60,
  boxrule=0.6pt,
  arc=2mm,
  left=3mm,right=3mm,top=2mm,bottom=2mm,
  width=\textwidth
}

\newtcolorbox{teacherbox}{
  enhanced,
  breakable,
  colback=yellow!25,
  colframe=yellow!55!black,
  boxrule=0.6pt,
  arc=2mm,
  left=2mm,right=2mm,top=1.2mm,bottom=1.2mm,
  width=0.9\textwidth,
  before skip=3pt,
  after skip=3pt
}

\newtcolorbox{studentbox}{
  enhanced,
  breakable,
  colback=green!20,
  colframe=green!45!black,
  boxrule=0.6pt,
  arc=2mm,
  left=2mm,right=2mm,top=1.2mm,bottom=1.2mm,
  width=0.9\textwidth,
  before skip=3pt,
  after skip=3pt
}

\newcommand{\Teacher}[1]{%
  \noindent\begin{flushleft}\begin{teacherbox}#1\end{teacherbox}\end{flushleft}%
}
\newcommand{\Student}[1]{%
  \noindent\begin{flushright}\begin{studentbox}#1\end{studentbox}\end{flushright}%
}

\newcolumntype{C}[1]{>{\centering\arraybackslash}p{#1}}

\usepackage{todonotes}
\usepackage{xcolor}

\newcommand{\mistral}{\texttt{Mistral-7B }}

\newcommand{\qwenseven}{\texttt{Qwen-7B }}

\newcommand{\qwenseventytwo}{\texttt{Qwen-72B }}

\newcommand{\modelname}{\textsc{Odqs }}

\newcommand{\unscaffolded}{\textsc{Unscaffolded }}

\newcommand{\scaffolded}{\textsc{Scaffolded }}

\newcommand{\static}{\textsc{Static }}

\newcommand{\prea}{\textsc{Pre-Assessment }}

\newcommand{\mida}{\textsc{Mid-Assessment }}

\newcommand{\self}{\textsc{Self-Odqs }}

\newcommand{\peer}{\textsc{Peer-Odqs }}

\title{Socratic Students: Training Language Models to Ask Better Questions for Reasoning}

\author{
 \textbf{Rajeev Bhatt Ambati \textsuperscript{1}$^*$},
 \textbf{Tianyi Niu \textsuperscript{1}$^*$},
 \textbf{Aashu Singh \textsuperscript{2}},\\
 \textbf{Shlok Mishra \textsuperscript{2}},
 \textbf{Snigdha Chaturvedi \textsuperscript{1}},  
 \textbf{Shashank Srivastava \textsuperscript{1}}
\\
\\
 \textsuperscript{1}UNC Chapel Hill,
 \textsuperscript{2}Meta
\\
}

\begin{document}
\maketitle
\def\thefootnote{*}
\footnotetext{Equal contribution. Correspondance: \href{mailto:email@domain}{ambati@cs.unc.edu}
 }

\begin{abstract}
Large language Models (LLMs) are usually used to answer questions, but many high-stakes applications (e.g., tutoring, clinical support) require the complementary skill of \textit{asking} questions: detecting missing information, requesting clarifications, and using them to solve tasks. We study this skill in reasoning-heavy domains where progress depends on inquiry rather than factual recall. We define an interactive protocol where a student model engages a stronger teacher under a small turn budget. After each teacher reply, we evaluate the student on the original task with Pass@k. We propose \textit{Outcome-Driven Question optimization Strategy} (\modelname), a training framework that learns a questioning policy from downstream task outcomes. At each turn, we sample multiple candidate questions; query the teacher with each, then score the student’s resulting performance. Using these scores, we train the student via supervised fine-tuning followed by Direct Preference Optimization (DPO), without any human labels. 
On GSM8K, HumanEval, and OpenCoder, \modelname produces large gains over interactive baselines, boosting Pass@5 by up to \textit{54.7\%} (absolute) on math and \textit{22.9\%} (absolute) on coding, and matching 
baseline performance in \textit{three fewer turns}. Thus, question asking can be explicitly trained from task outcomes, improving both accuracy and efficiency in interactive reasoning.

\end{abstract}

\section{Introduction}

The dominant paradigm for language models is reactive: present a prompt and receive a response. This works beautifully when the model has the information it needs. However, many real-world applications, such as educational tutoring \citep{avis,pan2024kwaiagentsgeneralizedinformationseekingagent,learnlmteam2025learnlmimprovinggeminilearning,mdagents} and medical assistance \citep{MediQ,li2025alfaaligningllmsask} require models to identify uncertainties, ask questions, and adapt to new information. For example, a diagnostic assistant must ask targeted questions before recommending treatment, or a tutor must ask probing questions to identify a student's misconceptions. In these settings, knowing \textit{what to ask} is the central bottleneck. In such \textit{dynamic interactions}, models fail not because they cannot generate answers, but because they ask the wrong questions, or none at all.

Recent work has explored interactive settings, including agents that ask clarifying questions \citep{aliannejadi2019clarifying,press2023measuring,yao2023react} and student–teacher setups where a stronger model guides a weaker one \citep{kendapadi-etal-2025-interact}. These approaches show that interaction helps, but a key gap remains: we lack a training signal that teaches a model \textit{which questions to ask}. Most methods rely on heuristics, scaffolds, or human judgments of question quality \citep{aliannejadi2019clarifying,yao2023react}. We argue that question quality should be judged not by style or surface semantics, but by utility: does the question improve the model’s ability to solve the task? Some work applies reward-based refinement for clarifying questions~\cite{andukuri2024star,srivastava-etal-2019-learning}, but previous work has not explored training questioning policies or interaction in reasoning tasks.

\begin{figure*}[ht]
  \centering
  \includegraphics[width=\linewidth, trim=0 9cm 0 0, clip]{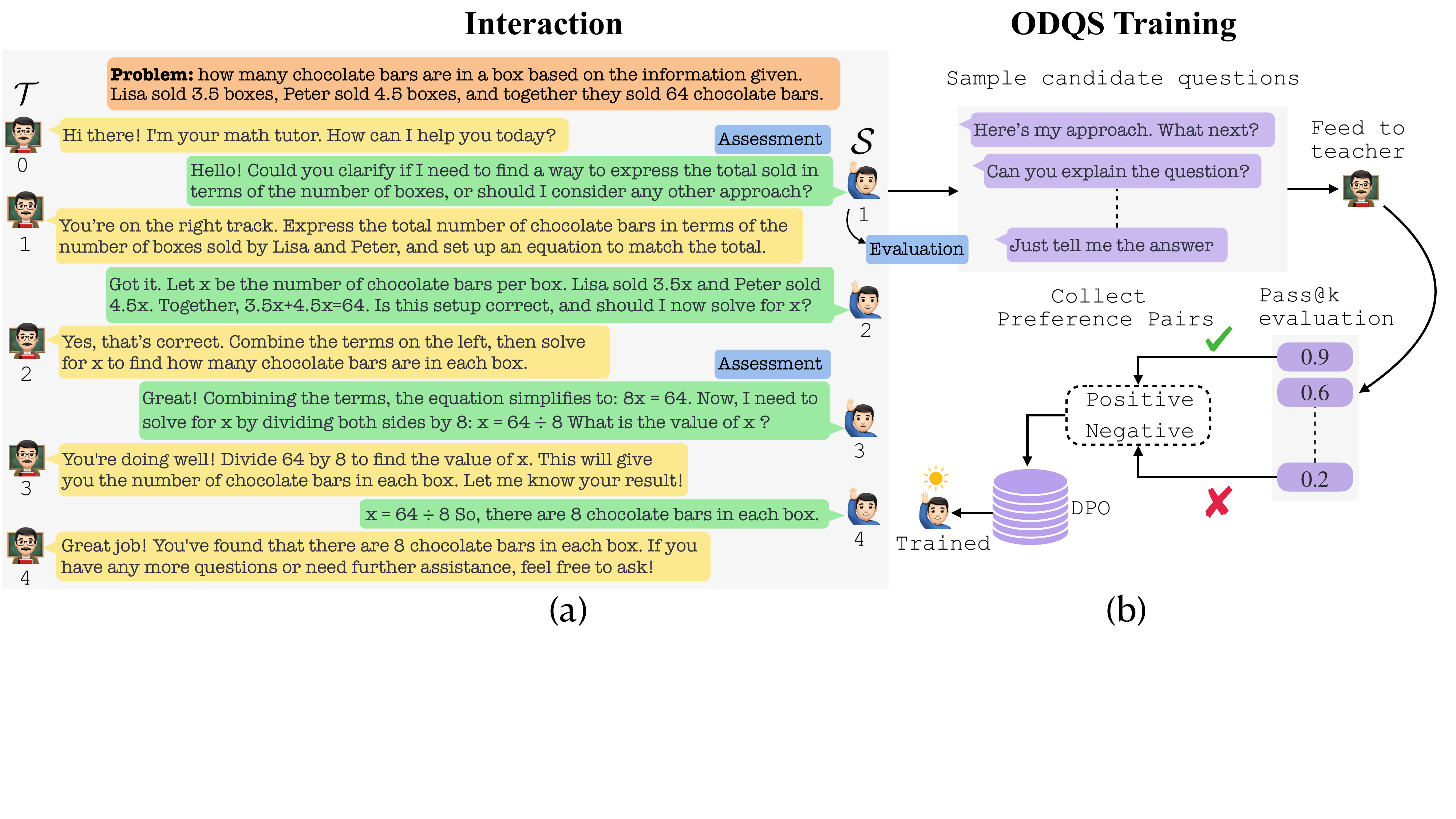}
    \caption{\textrm{Overview of our interactive learning framework.}
    (a) \textbf{Interaction:} A student model ($\mathcal{S}$) interacts with a stronger teacher model ($\mathcal{T}$) over $5$ turns of questions and responses. After each teacher response, $\mathcal{S}$'s performance on the problem is evaluated using \textit{Pass@k}.
    (b) \textbf{\modelname Training:} To train $\mathcal{S}$ to ask better questions, we collect preference pairs for DPO. At each turn, we sample $k$ candidate questions $\{q_1,\dots,q_k\}$. Each question is evaluated based on downstream \textit{Pass@k}, and preference pairs are formed accordingly. $\mathcal{S}$ is then trained on these pairs.
    \textbf{Assessment:} We explore injecting intermediate assessments, where the teacher can provide feedback on $\mathcal{S}$'s attempts on the problem. We explore 
    including assessments at turn~$0$ (\prea) or at turn~$2$ (\mida).
    }
  \label{fig:main_overview}
\end{figure*}

We focus on reasoning-intensive domains (math and code) and formalize the interaction via a student-teacher protocol: a student model $\mathcal{S}$ attempts a problem and is allowed to query a stronger teacher $\mathcal{T}$, that provides guidance but never the final answer. The student operates under a budget of questioning turns. After each teacher response, we evaluate whether $\mathcal{S}$ can now solve the original problem by sampling answer-only attempts and computing \textit{Pass@$k$}. This yields a clean operational definition of utility: a question is \textit{good} if and only if it increases downstream \textit{Pass@$k$}.

Building such agents presents three core challenges. The first challenge is \textit{search}, since the space of possible questions that can be posed in natural language is immense. The second challenge is \textit{supervision}, since there are no existing datasets of ground-truth best questions to ask.  
Current post-training methods such as SFT \citep{ouyang2022traininglanguagemodelsfollow} or DPO \citep{rafailov2024directpreferenceoptimizationlanguage} are primarily used for style, tone, and subjective quality of responses \citep{bai2022traininghelpfulharmlessassistant}, not information seeking. The third challenge is \textit{efficiency}, since in real deployments, interactions are expensive, so improvements must come from making substantial progress per turn, not simply adding more dialogue.


To bridge these gaps, we introduce two classes of questioning strategies. Our first approach, \textit{assessment-based questioning}, is a prompting strategy where the student attempts to solve the problem at the start (\prea) or mid-interaction (\mida). The teacher then assesses this attempt, and the student uses that feedback to ask the next question.
Our second, \textit{outcome-driven question optimization} (\modelname), is a supervised method that frames question asking as an outcome-labeled decision problem. At each turn, multiple candidate questions are generated. Each candidate is executed by querying the teacher, and the resulting trajectory is scored using the student’s downstream \textit{Pass@$k$} performance on the problem. The highest-scoring candidate is treated as a \textit{preferred} example, while lower-scoring candidates serve as contrasts. These preference pairs are then used to train a questioning policy. This approach enables models to bootstrap effective questioning behavior without human-labeled question quality.

On GSM8K (math) and HumanEval/OpenCoder (code), interactive baselines consistently improve \textit{Pass@k} over a static baseline, with absolute gains of at least $0.5$. \modelname also improves interaction efficiency: \modelname-trained students match the final performance of Unscaffolded interaction using three fewer turns. Overall, \modelname boosts \textit{Pass@5} by up to 54.7\% on math and 22.9\% on coding. We also find the optimal position for adding assessments, and analyze its impact on both domains. Our contributions are:
\begin{itemize}[itemsep=0pt,topsep=5pt,leftmargin=10pt]
    \item We develop \modelname, a DPO-based approach that trains models to ask questions based on reasoning outcomes rather than human judgments.
    \item We introduce assessment-based interaction strategies that provide targeted feedback to students to improve question asking and learning.
    \item We empirically show that asking good questions is a learnable skill that improves accuracy and interaction efficiency across models and benchmarks.
    \item We provide an in-depth analysis of interactive learning dynamics, explaining how improved progress per turn and earlier high-quality questions drive performance gains.

\end{itemize}

\section{Related Work}

\paragraph{General Interactive Learning:}
Early work on language-guided machine learning relied on single-turn instructions \citep{srivastava-etal-2017-joint, hancock-etal-2018-training, labutov-etal-2018-lia, arabshahi2020conversational}. To improve comprehension, researchers investigated active learning \citep{activelearning, tamkin2022activelearninghelpspretrained} and language-based clarification \citep{rao-daume-iii-2018-learning, srivastava-etal-2019-learning}. Recent work probes LLMs’ information-seeking abilities. \citet{bertolazzi-etal-2023-chatgpts} studied the 20-questions game using an information-theoretic lens, though its binary format limits extension to richer interactions. Other lines develop tool-using QA agents~\citep{pan2024kwaiagentsgeneralizedinformationseekingagent}, improve retrieval via query rephrasing \cite{deng2024rephraserespondletlarge}, and extend interactive learning to multimodal setups~\citep{avis}. Closely related to our work, \citet{kendapadi-etal-2025-interact} study teacher-led interactions in knowledge domains such as lyrics and news, primarily testing factual knowledge. In contrast, we focus on domains where reasoning rather than retrieval is the main challenge.

\paragraph{Application contexts:}
Interactive frameworks have been explored in high-stakes domains such as medicine and education. In clinical support, prior work studies adaptive collaborations among LLMs \citep{mdagents}, diagnostic-style reasoning frameworks \citep{chen-etal-2025-cod}, and uncertainty-triggered follow-ups \citep{MediQ,hu2024uncertaintythoughtsuncertaintyawareplanning}, as well as approaches based on static question sets or handcrafted attributes \citep{winston2024medical,li2025alfaaligningllmsask}. In education, LLM tutors have been evaluated for pedagogical quality \citep{tack2022aiteachertestmeasuring}, with datasets and post-training methods developed to improve tutoring behavior \citep{jurenka2024responsibledevelopmentgenerativeai,chevalier2024languagemodelssciencetutors,kwon-etal-2024-biped,dan2023educhatlargescalelanguagemodelbased,sonkar-etal-2023-class}. 

\section{Questioning Strategies}
\label{sec:methods}

We study whether a student model can improve its performance on reasoning tasks by asking helpful questions to a stronger teacher. Let $\mathcal{S}$ denote the student and $\mathcal{T}$ the teacher. The ultimate goal for the student is to solve a reasoning problem $\mathcal{P}$. For this, $\mathcal{S}$ interacts with $\mathcal{T}$ for a fixed budget of $N$ \textit{questioning turns} (we use $N=5$), as illustrated in Figure~\ref{fig:main_overview}. At each turn $i \in \{1\ldots N\}$, the student asks a single targeted question, and the teacher returns a clarification while being explicitly instructed not to reveal the final answer to $\mathcal{P}$. After each teacher utterance, we evaluate the student’s ability to solve $\mathcal{P}$ using the full interaction history so far: we sample $k$ direct answers from $\mathcal{S}$ and compute \textit{Pass@k} \citep{chen2021evaluatinglargelanguagemodels, wang2023selfconsistencyimproveschainthought}.

\subsection{Baseline methods}
\label{baselines}

\noindent \textbf{\static}: Our primary baseline is the \static baseline that corresponds to student model performance before any interactions, i.e., at turn~0.

We also consider two prompt-based interactive baselines: \unscaffolded and \scaffolded. 

\noindent \textbf{\unscaffolded}:
At each student utterance, $\mathcal{S}$ is asked to directly generate a question for $\mathcal{T}$ without any explicit scaffolding. In our experiments, we compare this method with our proposed strategies to measure how much performance improves from interaction alone, without any guidance about how to formulate questions.

\noindent \textbf{\scaffolded}:
In each turn, the student’s system prompt is augmented with a lightweight CoT-style self-reflection scaffold. Before posing a question, $\mathcal{S}$ is instructed to (A) summarize what is already known, (B) identify the missing sub-goals needed to solve the task, and (C) ask a follow-up question. This encourages targeted information seeking.

\unscaffolded and \scaffolded are instantiations of Algorithm~\ref{alg:vanilla} without any assessment ($t_{\text{assess}}=-1$). Their prompts are provided in Appendix~\ref{app:prompts_interaction}.

\subsection{Assessment-based Strategies}
Similar to human learners, we hypothesize that timely feedback can help LLM students calibrate their understanding and guide subsequent questioning. We examine whether a similar mechanism improves interactive learning for LLMs. We propose two prompting-based variants that insert a one-shot assessment at different points in the interaction. In an assessment, the student ($\mathcal{S}$) generates a candidate solution to $\mathcal{P}$, and the teacher ($\mathcal{T}$) provides feedback on its correctness and highlights errors in the student’s reasoning. This assessment is then appended to the conversation history, grounding all subsequent turns.

\noindent \textbf{\prea~:}
Here the assessment is introduced before the student’s first turn ($t_{\text{assess}}=0$). This allows the student to be explicitly grounded from the very beginning of the interaction. This corresponds to Algorithm~\ref{alg:vanilla} with $t_{assess} = 0$.

\noindent \textbf{\mida:}
Here the assessment occurs halfway through the interaction ($t_{\text{assess}}=2$ out of $N=5$ turns). Unlike pre-assessment, the student can leverage the accumulated dialogue from the first three turns. This corresponds to Algorithm~\ref{alg:vanilla} with $t_{\text{assess}}=2$.

\noindent See Appendix~\ref{app:prompts_assessment} for prompts used in assessments.

\subsection{Outcome-Driven Question-optimization Strategy (\modelname)}
\label{sec:modelname_strategy}
We frame question asking as a decision problem supervised by downstream task performance. Our approach first collects candidate questions and then fine-tunes students to ask better questions based on their impact on task success. At each student turn, we sample multiple candidate questions for the student to ask the teacher. For each candidate, we query the teacher, append the resulting teacher reply to the dialogue history, and evaluate the student by downstream \textit{Pass@k} on the original problem. 

We then form preference pairs by marking the highest-scoring candidate as \textit{chosen} and lower-scoring candidates as \textit{rejected}. From these, we build an SFT dataset $\mathcal{B}_{\mathrm{SFT}}$ (prompt, chosen) and a DPO dataset $\mathcal{B}_{\mathrm{DPO}}$ (prompt, chosen, rejected) tuples where the prompt is the same question-generation prompt used by the student. The student is fine-tuned with SFT followed by DPO. This procedure is summarized in Algorithm~\ref{alg:guided_sg}. We propose two variants of this strategy.

\noindent \textbf{\self:}
Here the set of candidate questions is sampled from the student model itself. 
The student learns from its own self-exploration, discovering which of its own questions are most effective.

\noindent \textbf{\peer:}
Here the set of candidate questions is sampled from 
a more capable peer model. 
The student learns its question asking strategy from these externally generated candidates. 

\section{Experiments}

\subsection{Datasets}
We evaluate our methods on two reasoning-intensive domains which require multi-step reasoning and benefit from interactive clarification: math and code. In the \textbf{Math} setting, a high-school–level arithmetic problem is given, and the student must reason step-by-step to compute the correct numerical answer. In the \textbf{Coding} setting, a natural-language problem description is provided, and the student must synthesize a correct Python program.

\textbf{Math:} We evaluate on GSM8K \citep{gsm8k}. For each student, we construct a subset of 2000 examples that are teacher-solvable but student-unsolvable. A problem is considered teacher-solvable if the teacher is able to solve the problem. It is student-unsolvable if the student's $Pass@5 < 0.5$ under five stochastic generations (temperature $0.3$) in the static (zero-shot) setting.

\noindent \textbf{Coding:} We evaluate on HumanEval \citep{humaneval} and OpenCoder \citep{huang-etal-2025-opencoder}. For each student, we construct a subset of 100 examples that are {teacher-solvable} but student-unsolvable. A problem is considered {teacher-solvable} if the teacher solution passes all the canonical unit tests. It is {student-unsolvable} if the student's $Pass@5 < 0.5$ under five stochastic generations (temperature $0.3$), with canonical unit tests.

Our \modelname strategy (\S~\ref{sec:modelname_strategy}) requires curated preference data of questions. Table~\ref{app:training_data} summarizes the number of collected examples used for training.

\subsection{Models}
We use \texttt{Qwen2.5-72B-Instruct} \citep{qwen2025qwen25technicalreport} as the teacher and two instruction-tuned models as student: (1) \texttt{Qwen2.5-7B-Instruct} \citep{qwen2025qwen25technicalreport}, and (2) \texttt{Mistral-7B-Instruct-v0.3} \citep{jiang2023mistral7b}. For \modelname-sampling strategies, we use \texttt{Qwen3-30B-A3B-Instruct-2507 }\cite{yang2025qwen3technicalreport} as the peer.

\subsection{Setup}
For tracking student's progress, we use a 5-turn interaction setup and evaluate the student’s performance before each student question ($t \in \{1,2,\dots,5\}$). 
At each evaluation step, we prompt the student model $\mathcal{S}$ for the final answer only and draw $k$ stochastic generations (temperature 0.3, max tokens 2048). \textit{Pass@k} \citep{chen2021evaluatinglargelanguagemodels, wang2023selfconsistencyimproveschainthought} is defined as the probability that any one of the $k$ generations matches the correct answer. We report the average \textit{Pass@k} at a given evaluation turn $t$. Details of \modelname-training are provided in Appendix~\ref{app:training_details}.


\section{Results}
In this section, we compare different questioning strategies. Our goal is not to compare models against each other, but to analyze how much each student \textit{improves over its own static baseline} through interaction. We therefore study learning dynamics via the trajectory of Pass@5 across turns $t \in \{1 \ldots 5\}$: effective strategies produce steadily increasing curves (with faster gains early gains being preferable), whereas flat trajectories indicate limited benefit from interaction. We organize our analysis around three research questions: (RQ1) Whether interactions improve performance over a static baseline?; (RQ2) Whether assessments help students to ask better questions; and (RQ3) Whether students can be trained to ask better questions? All results are shown in Figures~\ref{fig:rq1_results}, ~\ref{fig:rq2_results}, ~\ref{fig:rq3_results}, and include bootstrapped error bars, which are too small to be noticeable in most cases. Tabulated results and statistical significance tests are reported in Appendix~\ref{app:results}. All the improvements reported are in absolute numbers. A concise summary of the main findings is provided in Table~\ref{tab:summary_results}.

\begin{table}[!htb]
\centering
\caption{\textbf{Summary of main findings across research questions.}
See the referenced sections for details.}
\label{tab:summary_results}

\begin{tabularx}{\columnwidth}{p{0.6cm} X p{0.4cm}}

\toprule
\textbf{RQ} & \textbf{Main Findings} & \textbf{Sec.} \\
\midrule

\textbf{RQ1} &
Even simple student--teacher interactions yield gains
up to $+0.85$ in \textit{Pass@k}.
Scaffolding provides additional improvements, especially in early turns.
& \S~\ref{sec:rq1_results} \\

\textbf{RQ2} &
Assessment timing is domain-dependent:
Pre variant works best for math (up to $+24.7\%$),
while Mid is most effective for coding (up to $+12.8\%$).
& \S~\ref{sec:rq2_results} \\

\textbf{RQ3} &
Self and Peer variants yield the strongest learning curves
(improvements up to $+49.3\%$) and matches the static performance with \textit{three fewer turns}.
& \S~\ref{sec:rq3_results} \\

\bottomrule
\end{tabularx}
\end{table}

\subsection{RQ1: Do interactions improve performance over a static baseline?}
\label{sec:rq1_results}

\begin{figure*}[t]
  \centering
  \begin{subfigure}[b]{0.24\textwidth}
    \includegraphics[width=\textwidth]{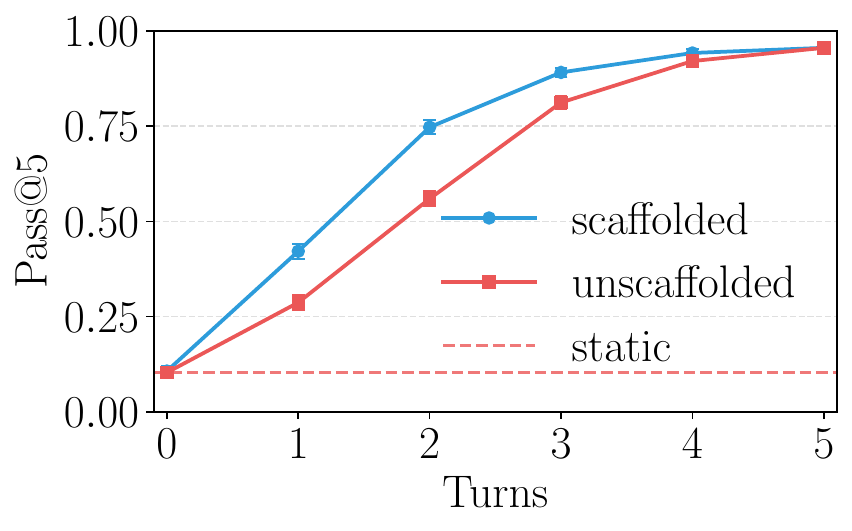}
    \caption{\qwenseven, Math}
    \label{fig:qwen7_math_pass5_rq1}
  \end{subfigure}
  \hfill
  \begin{subfigure}[b]{0.24\textwidth}
    \includegraphics[width=\textwidth]{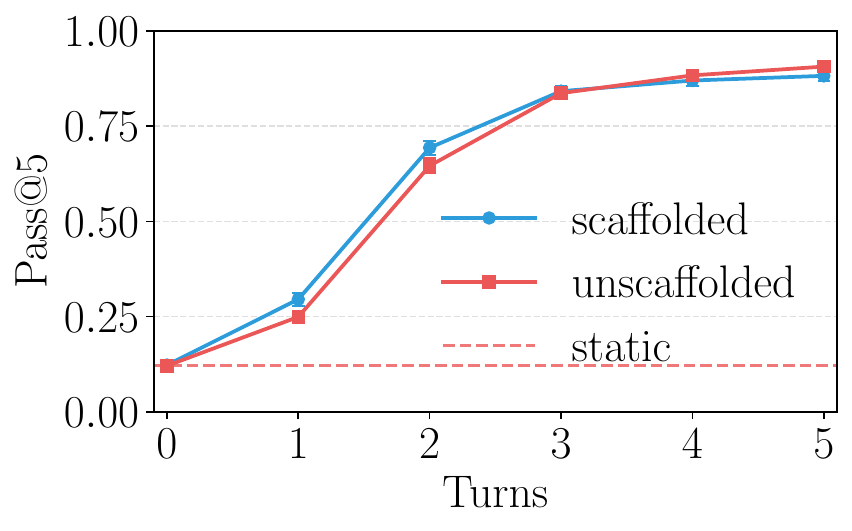}
    \caption{\mistral, Math}
    \label{fig:mistral7_math_pass5_rq1}
  \end{subfigure}
  \hfill
  \begin{subfigure}[b]{0.24\textwidth}
    \includegraphics[width=\textwidth]{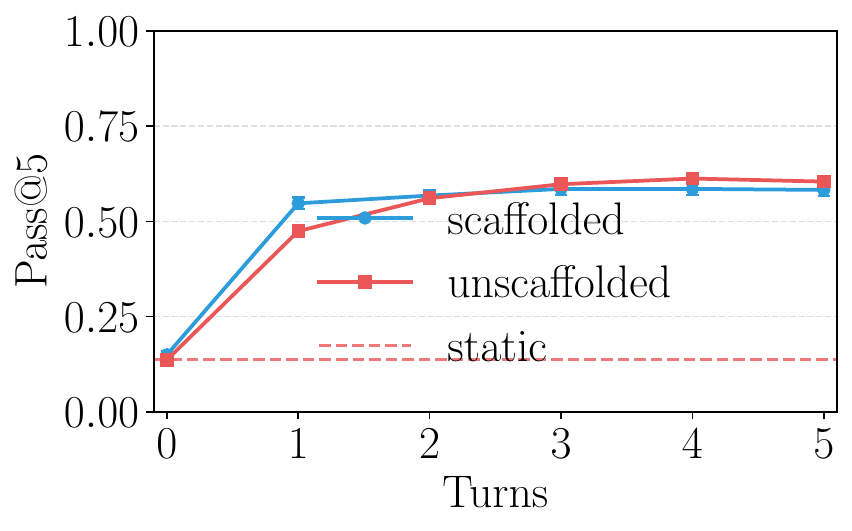}
    \caption{\qwenseven, Coding}
    \label{fig:qwen7_code_pass5_rq1}
  \end{subfigure}
  \hfill  
  \begin{subfigure}[b]{0.24\textwidth}
    \includegraphics[width=\textwidth]{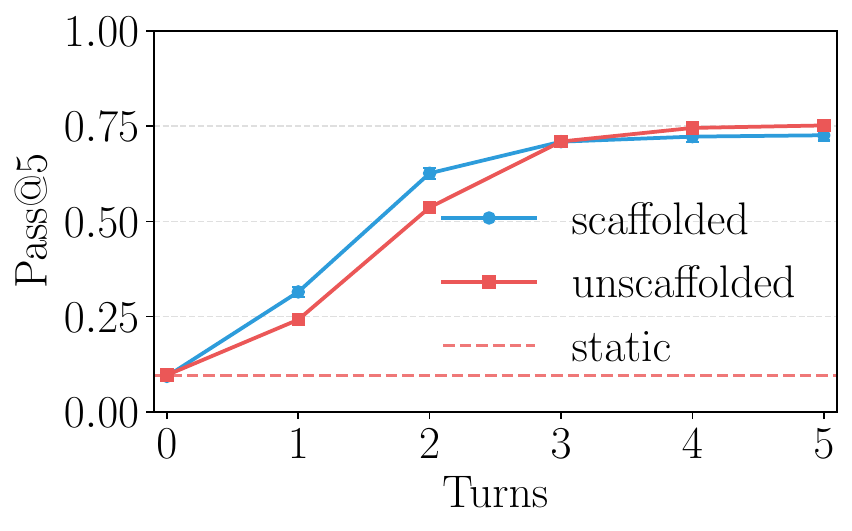}
    \caption{\mistral, Coding}
    \label{fig:mistral7_code_pass5_rq1}
  \end{subfigure}

\caption{
\textbf{Static vs Interactive Baselines} performance across 5 turns for two student models (\qwenseven and \mistral) on Math \textbf{(a, b)} and Coding \textbf{(c, d)}. \unscaffolded interaction already yields substantial gains for both models and domains. Adding \scaffolded further improves performance across turns.}

  \label{fig:rq1_results}
\end{figure*}

In this research question, we test whether even simple interaction helps. We compare the performance of the \unscaffolded and \scaffolded baselines with the \static baseline. Results for the two student models appear in Figures~\ref{fig:qwen7_math_pass5_rq1} and~\ref{fig:mistral7_math_pass5_rq1} (Math), and Figures~\ref{fig:qwen7_code_pass5_rq1} and~\ref{fig:mistral7_code_pass5_rq1} (Coding).

\paragraph{\unscaffolded interaction yields strong improvements.}
First, we compare the performance of \unscaffolded (red) with the \static baseline (dashed). We see that even with \unscaffolded interaction, on math, the performance of \qwenseven improves by $+0.85$ absolute \textit{Pass@5} over the \static baseline over the course of the interaction (5 turns). Similarly, the performance of \mistral improves by $+0.78$. Coding shows similar trends, with gains of $+0.47$ for \qwenseven and $+0.65$ for \mistral. Thus even the simplest interactive baseline significantly improves performance relative to the \static baseline.

\paragraph{\scaffolded helps most in early turns.}
Analyzing performance of \scaffolded (blue), we see that it provides similar gains over the \static baseline and consistent but smaller gains on top of \unscaffolded. On math, \scaffolded improves over \unscaffolded by an average of $+7\%$ (up to $+18.8\%$) for \qwenseven and by $+1\%$ on average (up to $+4.7\%$) for \mistral. On coding, it yields $+0.5\%$ on average (up to $+7.3\%$) for \qwenseven and $+1.8\%$ on average (up to $+8.9\%$) for \mistral. Across settings, these gains are most pronounced in early turns and saturate later.


\subsection{RQ2: Do assessments help students improve by asking better questions?}
\label{sec:rq2_results}

\begin{figure*}[t]
  \centering
  \begin{subfigure}[b]{0.24\textwidth}
    \includegraphics[width=\textwidth]{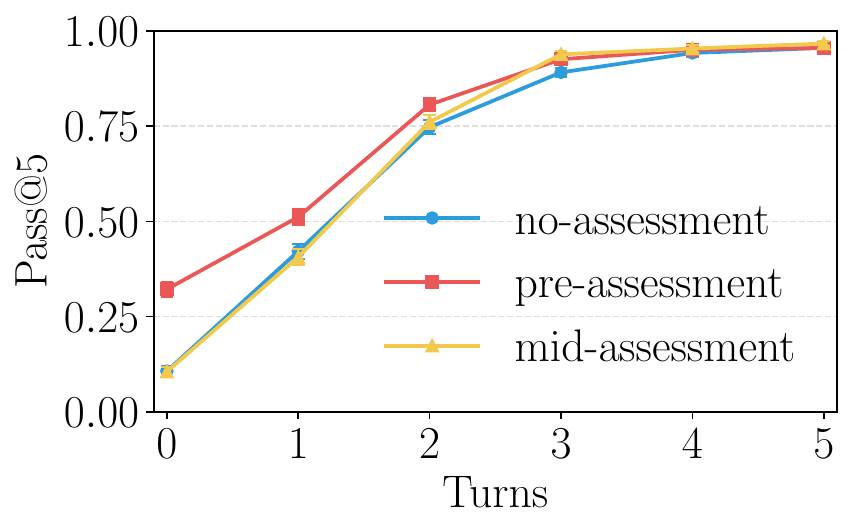}
    \caption{\qwenseven, Math}
    \label{fig:qwen7_math_pass5_rq2}
  \end{subfigure}
  \hfill
  \begin{subfigure}[b]{0.24\textwidth}
    \includegraphics[width=\textwidth]{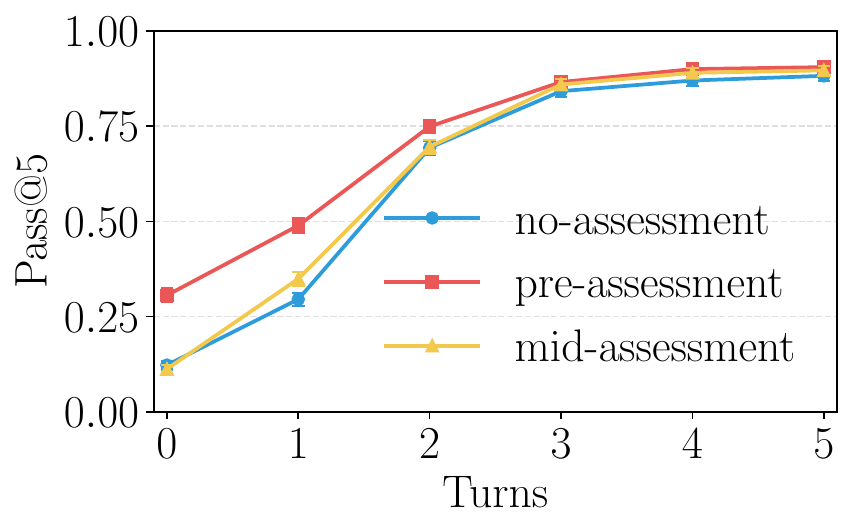}
    \caption{\mistral, Math}
    \label{fig:mistral7_math_pass5_rq2}
  \end{subfigure}
  \hfill  
  \begin{subfigure}[b]{0.24\textwidth}
    \includegraphics[width=\textwidth]{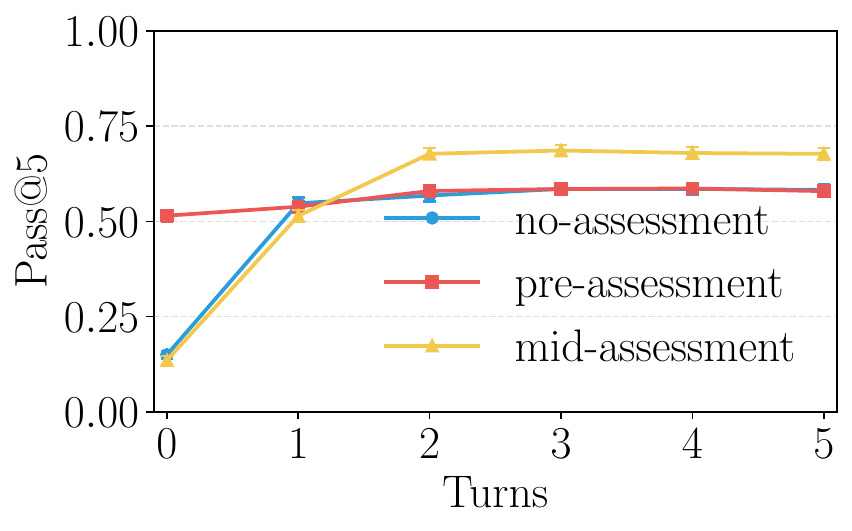}
    \caption{\qwenseven, Coding}
    \label{fig:qwen7_code_pass5_rq2}
  \end{subfigure}
  \hfill  
  \begin{subfigure}[b]{0.24\textwidth}
    \includegraphics[width=\textwidth]{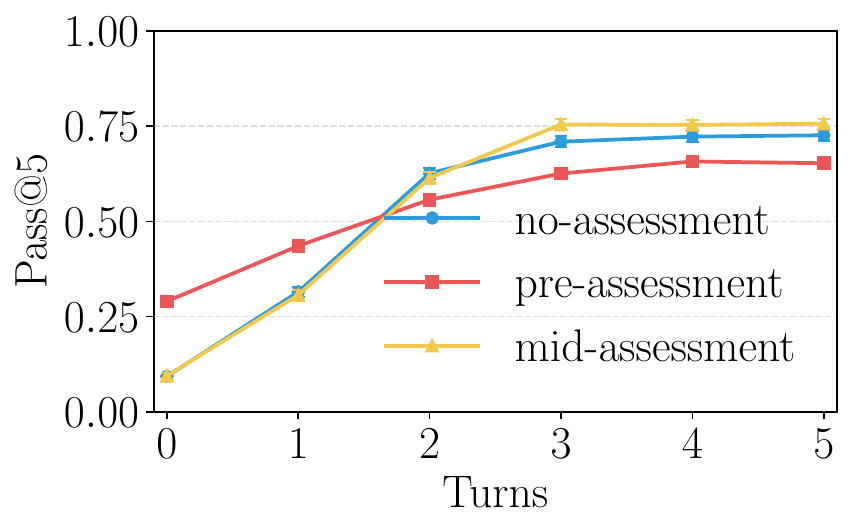}
    \caption{\mistral, Coding}
    \label{fig:mistral7_code_pass5_rq2}
  \end{subfigure}

\caption{
\textbf{Assessment Strategies} performance across 5 turns for two student models (\qwenseven and \mistral) on Math \textbf{(a, b)} and Coding \textbf{(c, d)}. Since \scaffolded improves performance in RQ1, we apply it by default. \prea provides the largest boost for math, whereas \mida is most effective for coding.}

  \label{fig:rq2_results}
\end{figure*}

In this research question, we examine whether providing explicit assessments helps students ask better questions, and how the timing of such assessments affects learning. Since Scaffolding improves early-turn learning in RQ1, we apply it by default. We compare the two assessment variants, \prea and \mida, with no-assessment, i.e. not providing an assessment. Results for both models are shown in Figures~\ref{fig:qwen7_math_pass5_rq2}, \ref{fig:mistral7_math_pass5_rq2} (Math), and Figures~\ref{fig:qwen7_code_pass5_rq2}, \ref{fig:mistral7_code_pass5_rq2} (Coding).



\paragraph{For Math, \prea outperforms \mida} We first compare \prea (red) and \mida (yellow) against no-assessment (blue). As expected, models immediately benefit from injected feedback. At turn~$0$, \prea improves over no-assessment by $+21.5\%$ for \qwenseven and $+18.3\%$ for \mistral. This early gain persists across subsequent turns: relative to no-assessment, \prea is better by $+6.8\%$ on average for \qwenseven (up to $+21.5\%$) and by $+8.5\%$ on average for \mistral (up to $+19.4\%$). In contrast, \mida yields only marginal improvements over no-assessment.

When comparing the two assessment positions directly, we observe no significant differences at later turns. However, \prea remains preferable because it introduces a performance boost from the very first turn. Moreover, \prea matches the final performance of the no-assessment setting using \textit{one fewer turn}, indicating higher efficiency under a fixed interaction budget.




\paragraph{For Coding, \mida outperforms \prea} At turn~$2$, \mida provides a strong boost, improving over no-assessment by $+10.9\%$ for \qwenseven. Across turns~$2$--$5$, \mida continues to outperform no-assessment by $+10.0\%$ on average for \qwenseven (up to $+10.9\%$) and by $+2.4\%$ on average for \mistral (up to $+4.5\%$). While \prea yields an initial improvement at turn~$0$, its benefits diminish quickly and remain marginal thereafter.

Crucially, \mida surpasses \prea in later turns: across turns~$2$--$5$, \mida outperforms \prea by $+9.7\%$ on average for \qwenseven (up to $+10.1\%$) and by $+9.6\%$ for \mistral (up to $+12.8\%$). Additionally, \mida reaches the same performance as no-assessment in \textit{two fewer turns}, demonstrating superior efficiency.

The \textit{optimal position} of assessment is domain-dependent. Math problems are stepwise and incremental, making early grounding (\prea) especially valuable. Coding problems, in contrast, are more divergent. An incorrect early direction can derail multiple subsequent turns. A delayed assessment (\mida) allows the model to first explore solution paths, then correct course once sufficient feedback is accumulated. These trends are examined in detail in \S~\ref{sec:analysis}.

\subsection{RQ3: Can we train students to ask better questions?}
\label{sec:rq3_results}

\begin{figure*}[t]
  \centering
  \begin{subfigure}[b]{0.24\textwidth}
    \includegraphics[width=\textwidth]{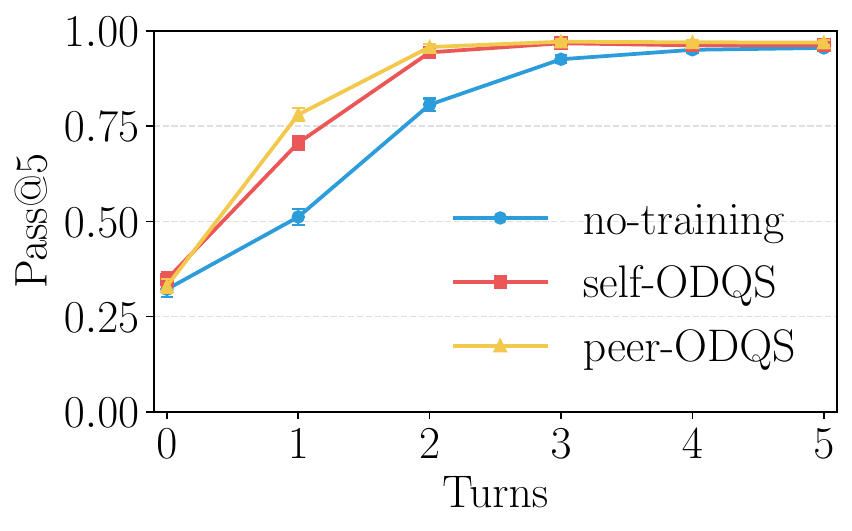}
    \caption{\qwenseven, Math}
    \label{fig:qwen7_math_pass5_rq3}
  \end{subfigure}
  \hfill
  \begin{subfigure}[b]{0.24\textwidth}
    \includegraphics[width=\textwidth]{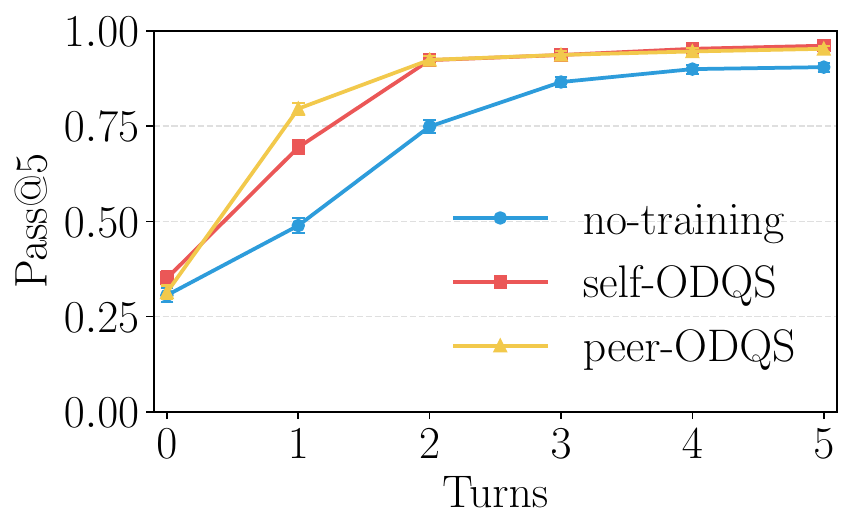}
    \caption{\mistral, Math}
    \label{fig:mistral7_math_pass5_rq3}
  \end{subfigure}
  \hfill  
  \begin{subfigure}[b]{0.24\textwidth}
    \includegraphics[width=\textwidth]{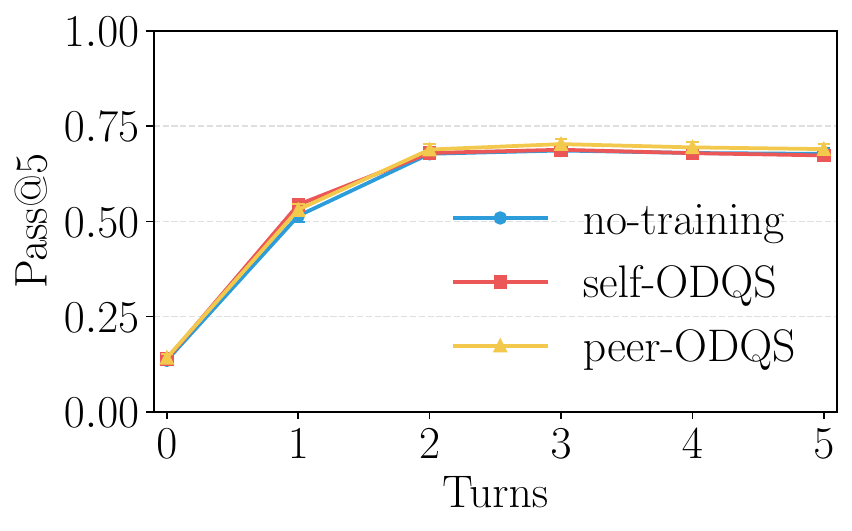}
    \caption{\qwenseven, Coding}
    \label{fig:qwen7_code_pass5_rq3}
  \end{subfigure}
  \hfill  
  \begin{subfigure}[b]{0.24\textwidth}
    \includegraphics[width=\textwidth]{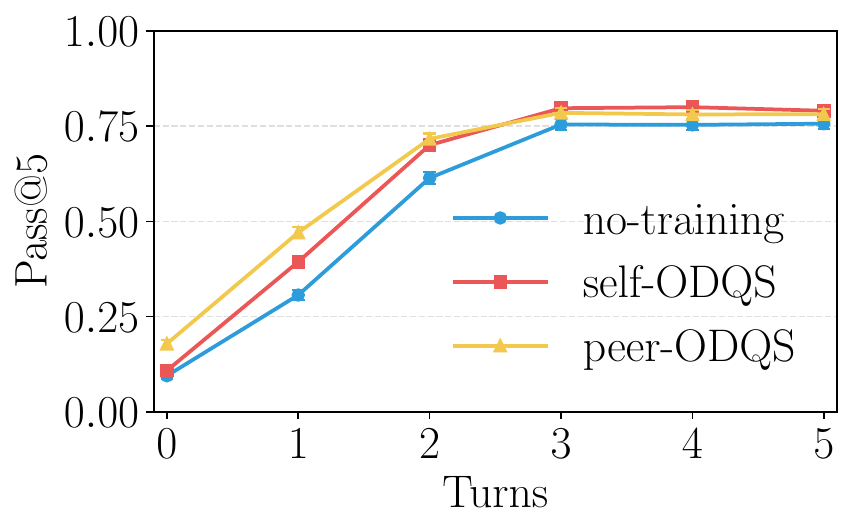}
    \caption{\mistral, Coding}
    \label{fig:mistral7_code_pass5_rq3}
  \end{subfigure}

\caption{
\textbf{\modelname Strategies} performance across 5 turns for two student models (\qwenseven and \mistral) on Math \textbf{(a, b)} and Coding \textbf{(c, d)}. Since assessment was found to be effective in RQ2, we apply \prea for Math and \mida for coding by default. Both \self and \peer outperform all other strategies and require \textbf{three fewer turns} to match the performance obtained without any training.}

  \label{fig:rq3_results}
\end{figure*}

In this research question, we evaluate whether students can be explicitly trained to ask better questions. Since assessments were shown to be beneficial in RQ2, we apply them by default: \prea for Math and \mida for Coding. We compare \self (red) and \peer (yellow) against no-training (blue), where no-training corresponds to \prea for Math and \mida for Coding. Results for both models are shown in Figures~\ref{fig:qwen7_math_pass5_rq3}, \ref{fig:mistral7_math_pass5_rq3} (Math), and Figures~\ref{fig:qwen7_code_pass5_rq3}, \ref{fig:mistral7_code_pass5_rq3} (Coding).


\paragraph{\modelname Training Strategy performs the best and is the most efficient.} On Math, both \self and \peer substantially outperform no-training across all turns. For \qwenseven, \self improves over no-training by $+7.0\%$ on average (up to $+19.4\%$), while \peer improves by $+8.5\%$ on average (up to $+26.8\%$). For \mistral, \self yields an average gain of $+10.1\%$ (up to $+20.5\%$), and \peer yields $+10.9\%$ on average (up to $+30.7\%$).



A similar trend holds for Coding. For \qwenseven, \self improves over no-training by $+0.6\%$ on average (up to $+3.0\%$), while \peer improves by $+1.4\%$ on average (up to $+1.7\%$). For \mistral, \self yields an average gain of $+5.2\%$ (up to $+8.7\%$), and \peer improves by $+7.2\%$ on average (up to $+16.4\%$).

Beyond \textit{Pass@5} gains, both \modelname variants are also more interaction-efficient: for both models and domains, they reach the same final performance as the assessment strategy using \textit{three fewer turns}. While \peer is slightly stronger at later turns, it incurs higher cost due to the need for a \textit{peer} model, making \self a more efficient alternative in practice.


\section{Analysis}
\label{sec:analysis}

\begin{figure}[!htb]
  \centering
    \includegraphics[width=0.4\textwidth]{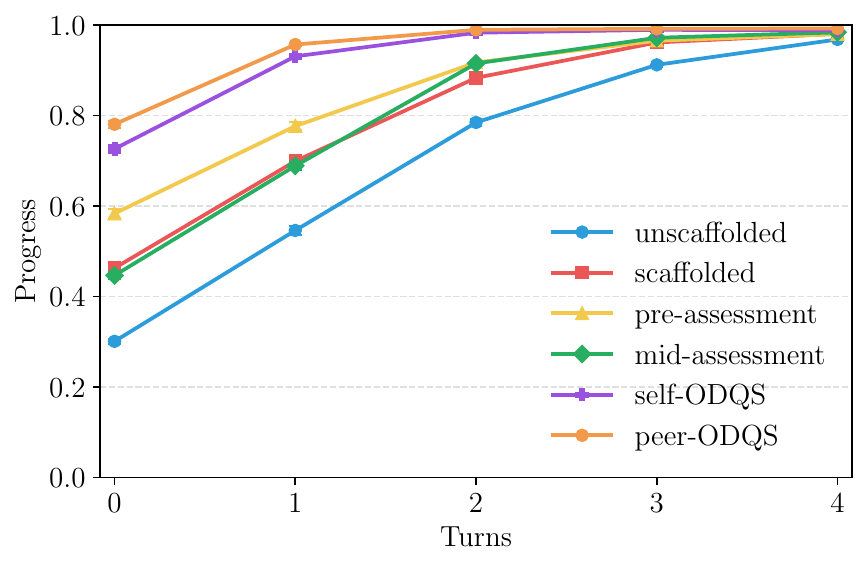}
    \caption{
    \textbf{Progress scores assigned by LLM-Judge across turns for \qwenseven on Math.} All guided methods help the student make more progress per turn compared to the \unscaffolded baseline. 
    Assessments produce sharp jumps at their insertion points, while \modelname training for both Self and Peer variants yield the highest progress throughout. 
    }
    \label{fig:qwen7_gsm8k_progress}
\end{figure}

\begin{figure}[!htb]
  \centering
    \includegraphics[width=0.4\textwidth]{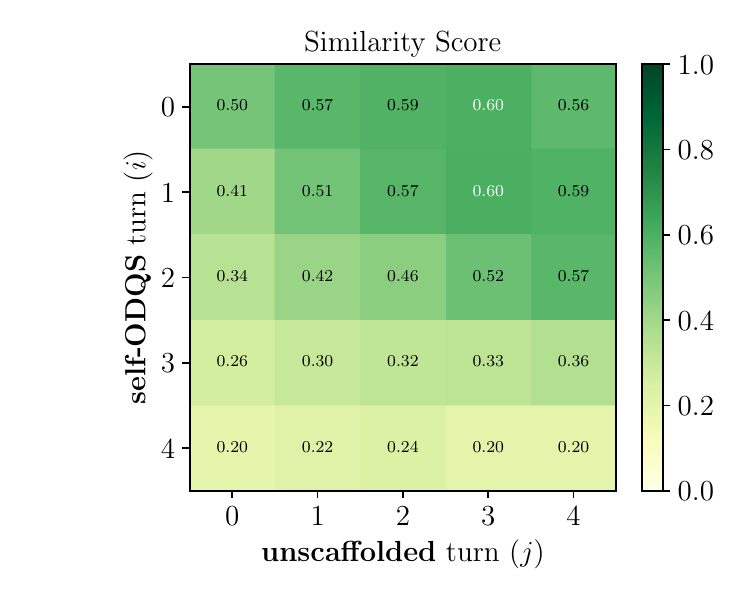}
    \caption{
    \textbf{Question similarity between \unscaffolded and \self} for \qwenseven questions Math. \self questions consistently resemble \textit{later-turn} \unscaffolded questions: 
    the \self question at Turn~$1$ is most similar to \unscaffolded questions from Turns~$2-5$, and the \self question at Turn~$2$ aligns with \unscaffolded Turns~$3-5$.
    This shows that \modelname training helps the model ask more advanced questions \textit{$2-3$ turns earlier} than the \unscaffolded baseline.
    }
    \label{fig:qwen7_gsm8k_similarity}
\end{figure}

In this section, we analyze \textit{why} and \textit{when} different questioning strategies outperform the \unscaffolded baseline, and examine the effect of assessment position on student performance.

\subsection{How does performance improve?}
We hypothesize that performance improves because better methods let the student (1) make more meaningful progress toward solving the problem at each turn, and (2) ask higher-quality questions earlier in the interaction. We empirically verify both claims. Here, we only analyze students' questions at turns $0-4$ since a question at the final turn ($5$) cannot be evaluated. 

\begin{figure}[!htb]
  \centering
  \begin{subfigure}{\linewidth}
    \centering
    \includegraphics[width=0.62\linewidth]{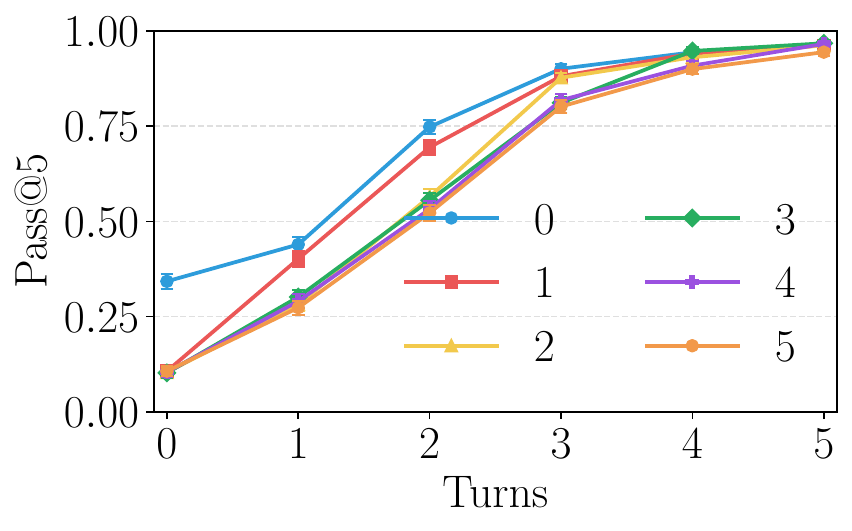}
    \caption{\qwenseven, Math}
    \label{fig:qwen7_gsm8k_assessment_pos}
  \end{subfigure}


  \begin{subfigure}{\linewidth}
    \centering
    \includegraphics[width=0.62\linewidth]{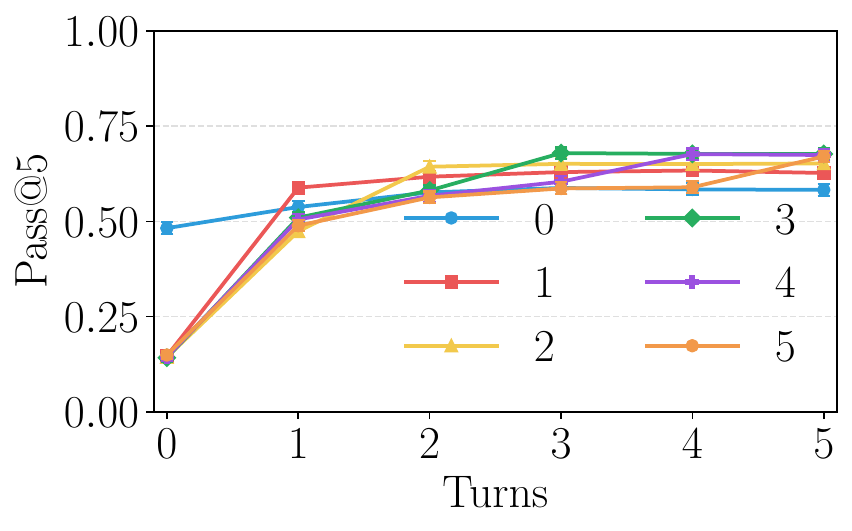}
    \caption{\qwenseven, Coding}
    \label{fig:qwen7_humaneval_opc_assessment_pos}
  \end{subfigure}

  \caption{
  \textbf{Impact of assessment position}.  
  \textit{Pass@5} for \qwenseven when inserting a single assessment at each turn ($0$--$5$).
  On math tasks (a), early assessment (Turn~$0$) provides the largest and most persistent improvements.  
  On coding tasks (b), the benefit of assessment is short-lived, with later assessments outperforming earlier ones.
  }
  \label{fig:qwen7_assessment_pos}
\end{figure}

\paragraph{Claim 1: Progress per turn.}
We use an \textit{LLM-as-a-Judge} evaluation: for each turn, we provide the problem, its gold solution, and the student's question and reasoning to \qwenseventytwo, which assigns a progress score in $[0,1]$ (prompts in Appendix~\ref{app:prompts_progress_analysis}). This score measures the student's \textit{reasoning progress}, not correctness (unlike \textit{Pass@5}).

Figure~\ref{fig:qwen7_gsm8k_progress} shows progress trajectories for \qwenseven on GSM8K.  
\scaffolded consistently increases progress per turn. Both assessments yield sharp improvements exactly at their insertion points (Turns~$1$ and $3$). The largest gains come from \modelname variants, Self and Peer, which outperform all other methods across turns.  
This verifies Claim~$1$.

\paragraph{Claim 2: Asking good questions earlier.}
We again use an LLM-as-a-Judge setup, using \qwenseventytwo, to compute the similarity between questions asked by \unscaffolded and \self students. Judge's prompts are shown in Appendix~\ref{app:prompts_similarity_analysis}. Figure~\ref{fig:qwen7_gsm8k_similarity} shows the heatmap comparing \self vs. \unscaffolded for Math. A darker color indicates higher similarity. Looking at the top-most row, we can see that \self questions at Turn~$0$ resemble \unscaffolded questions at Turns~$1-4$. Similarly, \self at Turn~$1$ resembles \unscaffolded at Turns~$2-4$. This pattern persists across all turns: \textbf{\self asks high-quality questions $1-3$ turns earlier}. This verifies Claim~$2$. We noticed similar observations for other methods in both Math and Coding domains. Example transcripts showing this behavior are provided in Appendix~\ref{app:examples}.

\paragraph{Conclusion:}
Both analyses show that better methods increase reasoning progress at each turn and enable the student to ask more advanced questions earlier, key drivers of their performance.

\subsection{Impact of assessment position}
To study how assessment timing affects performance, we place a single assessment at each turn from $0$ to $5$ and evaluate on both math and coding problems. Figure~\ref{fig:qwen7_assessment_pos} reports \textit{Pass@5} for \qwenseven across all assessment positions.

On \textbf{math}, assessment at Turn~$0$ performs best: it produces a large boost at Turn~$0$ that persists across the remaining turns.
On \textbf{coding}, improvement is transient: assessment at Turn~$0$ boosts Turn~$0$, but is soon overtaken by assessments at Turns~$1$ and $2$, and so on. Later assessments yield better performance because coding problems are not stepwise, early incorrect attempts often derail the solution, and a delayed assessment can correct accumulated errors. This explains the earlier empirical result:  
\begin{itemize}[itemsep=0pt,topsep=5pt]
    \item \textbf{Math}: \prea (Turn $0$) is best.  
    \item \textbf{Coding}: \mida (Turn $2$) is best.
\end{itemize}

\section{Conclusion}
We investigated whether LLMs can be prompted and trained to behave as effective interactive learners. Our results show that even a simple \unscaffolded student--teacher interaction yields substantial gains, with absolute improvements in \textit{Pass@k} often exceeding $+0.5$ and reaching as high as $+0.85$. \scaffolded guidance further amplifies these gains. We also found that assessment timing matters: \prea is highly beneficial for math tasks, whereas \mida is most effective for coding tasks, reflecting fundamental differences in their problem structures (stepwise vs.\ non-stepwise).

Beyond prompting strategies, we demonstrated that the skill of asking good questions can be explicitly trained. Both \self and \peer substantially improve students' questioning policies, leading to stronger downstream performance and greater learning efficiency, often achieving the same performance as \unscaffolded students with \textit{three fewer turns}. Our analysis further showed that stronger methods help models make more progress per turn and ask higher-quality questions earlier in the trajectory.

Overall, these findings indicate that LLMs can move beyond static information retrieval and operate as adaptive, interactive learners. A key challenge ahead is the reliance on brittle, in-context memory: developing more robust mechanisms for retaining and updating knowledge across turns is an important next step toward building truly adaptive learning agents.

\section*{Limitations}
Our framework assumes access to a reliable teacher that can (i) answer questions correctly and (ii) provide meaningful feedback during assessments. Performance may degrade if the teacher is noisy, biased, or weaker than the student, which limits applicability in low-resource or fully autonomous settings.

We define question quality indirectly through downstream task performance. While effective, this may conflate good questioning with favorable answer sampling, and does not explicitly capture pedagogical properties such as clarity, minimality, or interpretability of questions.

We evaluate our methods on math (GSM8K) and coding (HumanEval/OPC) tasks, which capture structured reasoning but do not cover broader interactive settings such as open-ended dialogue, long-horizon planning, or multimodal environments. The effectiveness of guided questioning in these domains remains an open question.

We study interactions under a small, fixed turn budget. Longer or adaptive interactions may exhibit different dynamics, and the efficiency gains observed here may not directly extrapolate to settings with unbounded or variable-length conversations.

Outcome-driven training requires executing multiple candidate questions per turn and evaluating them via teacher interaction, which can be computationally expensive. While effective, this may limit scalability to very large datasets or models without further optimization.

All learning occurs within the context window or via parameter updates. The framework does not address long-term memory, retrieval, or knowledge consolidation across tasks, which are important for truly adaptive interactive agents.

\bibliography{custom}

\appendix

\section{Methods}
\label{app:methods}
Algorithm~\ref{alg:vanilla} describes the \unscaffolded and \scaffolded based student–teacher interaction framework. Algorithm~\ref{alg:guided_sg} presents the self-training and peer-training approaches, including data collection, DPO training, and subsequent interaction-based evaluation.

\begin{algorithm*}[!htb]
\caption{Student--Teacher Interaction (\unscaffolded/\scaffolded, Assessment at Turn $t_{assess}$)}
\label{alg:vanilla}
\begin{algorithmic}[1]
\Require Dataset $\mathcal{D}=\{(x_i,y_i)\}_{i=1}^N$; Student $\mathcal{S}$; Teacher $\mathcal{T}$; total turns $N=5$; eval samples $k$; $\text{mode}\in\{\textsc{Unscaffolded},\textsc{Scaffolded}\}$; assessment turn $t_{assess} \in \{0,\dots,N\}$
\Function{Interact}{$\mathcal{D}, \mathcal{S}, \mathcal{T}, N, k, t_{assess}$}
  \For{\textbf{each} $(x,y)$ in $\mathcal{D}$}
    \State $H \gets$ \Call{InitHistory}{$x$}    \Comment{turn-0 initialized with a teacher message}
    \For{$t = 1$ \textbf{to} $N + 1$}  \Comment{Each iteration = Teacher then Student (2 turns)}

      \If{$t = t_{assess}$}
        \State $a \gets$ \Call{Assess}{$\mathcal{S}, H, y$};\; $H \gets H \cup \{a\}$ \Comment{Assessment turn}
      \EndIf

      \If{$(t\mod 2) == 1$} \Comment{Student at the odd turn}
      \State \Call{EvaluateStudent}{$\mathcal{S}, H, k, y$} \Comment{$k$ direct-answer generations}
      \State $m \gets$ \Call{Student}{$\mathcal{S}, H$}
      \EndIf

      \If{$(t\mod 2) == 0$} \Comment{Teacher at the even turn}
      \State $m \gets$ \Call{Teacher}{$\mathcal{T}, H$}
      \EndIf

      \State $H \gets H \cup \{m\}$
    \EndFor
  \EndFor
\EndFunction
\end{algorithmic}
\end{algorithm*}

\begin{algorithm*}[!htb]
\caption{\modelname Training: Self or Peer}
\label{alg:guided_sg}
\begin{algorithmic}[1]
\Require Dataset $\mathcal{D}=\{(x_i,y_i)\}_{i=1}^N$; Student $\mathcal{S}$; Teacher $\mathcal{T}$; total turns $N=5$; eval samples $k$; candidates $m$
\State \textbf{Buffers:} $\mathcal{B}_{\mathrm{SFT}} \gets \emptyset$,\; $\mathcal{B}_{\mathrm{DPO}} \gets \emptyset$
\Function{\textsc{Odqs}Interact}{$\mathcal{D}, \mathcal{S}, \mathcal{T}, \mathcal{G}, N, k, m$}
  \For{\textbf{each} $(x,y)$ in $\mathcal{D}$}
    \State $H \gets$ \Call{InitHistory}{$x$}    \Comment{Turn-0 initialized with a teacher message}
    \For{$t = 1$ \textbf{to} $N // 2$} \Comment{Each iteration = Teacher then Student (2 turns)}
      \State $Q \gets \emptyset$
      \For{$j=1$ \textbf{to} $m$}
        \State $q_j \gets$ \Call{Guide}{$\mathcal{G}, H$} \Comment{$G$ proposes next question}
        \State $h_j \gets$ \Call{Teacher}{$\mathcal{T}, H \cup \{q_j\}$} \Comment{Teacher responses with $q_j$}
        \State $s_j, \hat{A}_j \gets$ \Call{EvaluateStudent}{$\mathcal{S}, H \cup \{q_j,h_j\}, k, y$}
        \State $Q \gets Q \cup \{(q_j,h_j,s_j,\hat{A}_j)\}$
      \EndFor
      \State $j^\star \gets \arg\max_j s_j$ \Comment{Pick the best question based on \textit{Pass@k}}
      \State $q^\star,h^\star,s^\star,\hat{A}^\star \gets (q_{j^\star},h_{j^\star},s_{j^\star},\hat{A}_{j^\star})$
      \State $H_{\text{prev}} \gets H$;\; $H \gets H \cup \{q^\star,h^\star\}$ \Comment{Update history with only the best pair}
      \State $\mathcal{B}_{\mathrm{SFT}} \gets \mathcal{B}_{\mathrm{SFT}} \cup \{(H_{\text{prev}}, q^\star, h^\star, \hat{A}^\star)\}$ \Comment{Chosen for SFT}
      \For{\textbf{each} $(q_j,h_j,s_j,\hat{A}_j) \in Q$ \textbf{with} $j \neq j^\star$}
        \State $\mathcal{B}_{\mathrm{DPO}} \gets \mathcal{B}_{\mathrm{DPO}} \cup \{(H_{\text{prev}}, q^\star \succ q_j)\}$ \Comment{Chosen-Rejected for DPO}
      \EndFor
    \EndFor
  \EndFor
  \State $\mathcal{S}^* \gets \Call{TrainSFT}{\mathcal{S}, \mathcal{B}_{\mathrm{SFT}}}$ \Comment{SFT on chosen}
  \State $\mathcal{S}^* \gets \Call{TrainDPO}{\mathcal{S}^*, \mathcal{B}_{\mathrm{DPO}}}$   \Comment{DPO on preference pairs}
  \State \Call{Interact}{$\mathcal{D}, \mathcal{S}^*, \mathcal{T}, T_{\max}, k$, \textsc{Scaffolded}} \Comment{Run interact with the trained student}
\EndFunction
\end{algorithmic}
\end{algorithm*}




\section{Datasets}
\begin{table}[!htb]
\centering
\caption{\textbf{\modelname training data sizes.} Number of interaction logs used for DPO, by model, domain, and guidance type. \peer data use Qwen3-30B-A3B-Instruct-2507 \citep{yang2025qwen3technicalreport} as the peer.}
\label{app:training_data}
\begin{tabular}{l l|c|c}
\toprule
\textbf{Task} & \textbf{Guidance} & \multicolumn{1}{c|}{\textbf{\qwenseven}} & \multicolumn{1}{c}{\textbf{\mistral}} \\
\midrule
\multirow{2}{*}{Math} 
& Self  & 3617 & 5219 \\
& Peer  & 4447  & 6495 \\
\midrule
\multirow{2}{*}{Coding} 
& Self  & 3958  & 5970 \\
& Peer  & 6495 & 7176 \\
\bottomrule
\end{tabular}
\end{table}

\section{Setup}
\label{app:training_details}
For DPO training we use QLoRA \citep{dettmers2023qloraefficientfinetuningquantized} with 4-bit quantization, $r{=}16$, $\alpha{=}32$, dropout $0.05$, batch size 4, AdamW ($\beta_1{=}0.9,\beta_2{=}0.95$), learning rate $5{\times}10^{-6}$, cosine decay, 5 epochs on NVIDIA RTX A6000 (48\,GB) with early stopping.

\section{Results}
\label{app:results}


This section provides tabulated performance improvements and statistical significance test results. 
For math, results are reported for \qwenseven in Table~\ref{app:all_results_math_qwen7} and for \mistral in Table~\ref{app:all_results_math_mistral7}. 
For coding, results are reported for \qwenseven in Table~\ref{app:all_results_coding_qwen7} and for \mistral in Table~\ref{app:all_results_coding_mistral7}. 

Statistical significance tests comparing methods are summarized by research question: 
Table~\ref{app:rq1_significance} for RQ1, 
Table~\ref{app:rq2_significance} for RQ2, and 
Table~\ref{app:rq3_significance} for RQ3.


\begin{table*}[!htb]
\centering
\small
\caption{\textbf{Turn-wise Pass@5 performance for \qwenseven on Math}. Mean, standard deviation ($\mathrm{std}$) are displayed as $\text{mean}_{(\text{std})}$.}
\label{app:all_results_math_qwen7}
\begin{tabular}{l|cccccc}
\toprule
\textbf{Method} & \textbf{$0$} & \textbf{$1$} & \textbf{$2$} & \textbf{$3$} & \textbf{$4$} & \textbf{$5$} \\
\midrule
\unscaffolded
& $0.103_{(0.012)}$
& $0.287_{(0.018)}$
& $0.559_{(0.020)}$
& $0.812_{(0.016)}$
& $0.921_{(0.011)}$
& $0.955_{(0.008)}$ \\

\scaffolded
& $0.107_{(0.013)}$
& $0.422_{(0.020)}$
& $0.747_{(0.018)}$
& $0.891_{(0.013)}$
& $0.942_{(0.010)}$
& $0.955_{(0.008)}$ \\

\midrule
\prea
& $0.322_{(0.019)}$
& $0.512_{(0.020)}$
& $0.806_{(0.016)}$
& $0.925_{(0.011)}$
& $0.950_{(0.009)}$
& $0.955_{(0.009)}$ \\

\mida
& $0.106_{(0.012)}$
& $0.406_{(0.020)}$
& $0.761_{(0.017)}$
& $0.938_{(0.010)}$
& $0.954_{(0.009)}$
& $0.967_{(0.007)}$ \\

\midrule
\self
& $0.348_{(0.019)}$
& $0.706_{(0.019)}$
& $0.944_{(0.010)}$
& $0.967_{(0.007)}$
& $0.962_{(0.008)}$
& $0.962_{(0.008)}$ \\

\peer
& $0.330_{(0.019)}$
& $0.780_{(0.017)}$
& $0.958_{(0.008)}$
& $0.971_{(0.007)}$
& $0.970_{(0.007)}$
& $0.969_{(0.007)}$ \\
\bottomrule
\end{tabular}
\end{table*}

\begin{table*}[!htb]
\centering
\small
\caption{\textbf{Turn-wise Pass@5 performance for \mistral on Math}. Mean, standard deviation ($\mathrm{std}$) are displayed as $\text{mean}_{(\text{std})}$.}
\label{app:all_results_math_mistral7}
\begin{tabular}{l|cccccc}
\toprule
\textbf{Method} & \textbf{$0$} & \textbf{$1$} & \textbf{$2$} & \textbf{$3$} & \textbf{$4$} & \textbf{$5$} \\
\midrule
\unscaffolded
& $0.121_{(0.012)}$
& $0.249_{(0.016)}$
& $0.646_{(0.019)}$
& $0.837_{(0.015)}$
& $0.883_{(0.013)}$
& $0.906_{(0.012)}$ \\

\scaffolded
& $0.123_{(0.012)}$
& $0.295_{(0.017)}$
& $0.693_{(0.018)}$
& $0.842_{(0.014)}$
& $0.870_{(0.013)}$
& $0.882_{(0.013)}$ \\

\midrule
\prea
& $0.306_{(0.018)}$
& $0.489_{(0.019)}$
& $0.749_{(0.017)}$
& $0.866_{(0.013)}$
& $0.900_{(0.012)}$
& $0.905_{(0.012)}$ \\

\mida
& $0.113_{(0.011)}$
& $0.349_{(0.018)}$
& $0.696_{(0.018)}$
& $0.859_{(0.014)}$
& $0.890_{(0.013)}$
& $0.896_{(0.012)}$ \\

\midrule
\self
& $0.351_{(0.019)}$
& $0.694_{(0.018)}$
& $0.923_{(0.011)}$
& $0.937_{(0.010)}$
& $0.953_{(0.008)}$
& $0.961_{(0.008)}$ \\

\peer
& $0.314_{(0.018)}$
& $0.796_{(0.016)}$
& $0.924_{(0.011)}$
& $0.937_{(0.010)}$
& $0.946_{(0.009)}$
& $0.953_{(0.008)}$ \\
\bottomrule
\end{tabular}
\end{table*}

\begin{table*}[!htb]
\centering
\small
\caption{\textbf{Turn-wise Pass@5 performance for \qwenseven on Coding}. Mean, standard deviation ($\mathrm{std}$) are displayed as $\text{mean}_{(\text{std})}$.}
\label{app:all_results_coding_qwen7}
\begin{tabular}{l|cccccc}
\toprule
\textbf{Method} & \textbf{$0$} & \textbf{$1$} & \textbf{$2$} & \textbf{$3$} & \textbf{$4$} & \textbf{$5$} \\
\midrule
\unscaffolded
& $0.136_{(0.010)}$
& $0.474_{(0.015)}$
& $0.561_{(0.015)}$
& $0.598_{(0.015)}$
& $0.612_{(0.015)}$
& $0.604_{(0.015)}$ \\

\scaffolded
& $0.150_{(0.010)}$
& $0.547_{(0.015)}$
& $0.568_{(0.016)}$
& $0.585_{(0.016)}$
& $0.585_{(0.016)}$
& $0.583_{(0.016)}$ \\

\midrule
\prea
& $0.515_{(0.015)}$
& $0.539_{(0.016)}$
& $0.580_{(0.016)}$
& $0.585_{(0.015)}$
& $0.586_{(0.015)}$
& $0.579_{(0.016)}$ \\

\mida
& $0.135_{(0.010)}$
& $0.514_{(0.015)}$
& $0.677_{(0.015)}$
& $0.686_{(0.015)}$
& $0.679_{(0.015)}$
& $0.677_{(0.015)}$ \\

\midrule
\self
& $0.139_{(0.010)}$
& $0.544_{(0.015)}$
& $0.679_{(0.015)}$
& $0.688_{(0.015)}$
& $0.679_{(0.015)}$
& $0.673_{(0.015)}$ \\

\peer
& $0.142_{(0.010)}$
& $0.531_{(0.015)}$
& $0.689_{(0.015)}$
& $0.703_{(0.014)}$
& $0.694_{(0.014)}$
& $0.690_{(0.015)}$ \\
\bottomrule
\end{tabular}
\end{table*}

\begin{table*}[!htb]
\centering
\small
\caption{\textbf{Turn-wise Pass@5 performance for \mistral on Coding}. Mean, standard deviation ($\mathrm{std}$) are displayed as $\text{mean}_{(\text{std})}$.}
\label{app:all_results_coding_mistral7}
\begin{tabular}{l|cccccc}
\toprule
\textbf{Method} & \textbf{$0$} & \textbf{$1$} & \textbf{$2$} & \textbf{$3$} & \textbf{$4$} & \textbf{$5$} \\
\midrule
\unscaffolded
& $0.097_{(0.008)}$
& $0.242_{(0.012)}$
& $0.537_{(0.015)}$
& $0.710_{(0.014)}$
& $0.745_{(0.013)}$
& $0.752_{(0.013)}$ \\

\scaffolded
& $0.093_{(0.008)}$
& $0.315_{(0.014)}$
& $0.626_{(0.015)}$
& $0.709_{(0.014)}$
& $0.722_{(0.014)}$
& $0.726_{(0.014)}$ \\

\midrule
\prea
& $0.290_{(0.014)}$
& $0.436_{(0.015)}$
& $0.557_{(0.015)}$
& $0.626_{(0.015)}$
& $0.657_{(0.015)}$
& $0.653_{(0.015)}$ \\

\mida
& $0.095_{(0.008)}$
& $0.307_{(0.014)}$
& $0.614_{(0.015)}$
& $0.754_{(0.013)}$
& $0.753_{(0.013)}$
& $0.756_{(0.013)}$ \\

\midrule
\self
& $0.108_{(0.008)}$
& $0.394_{(0.015)}$
& $0.700_{(0.014)}$
& $0.797_{(0.013)}$
& $0.800_{(0.012)}$
& $0.790_{(0.013)}$ \\

\peer
& $0.179_{(0.011)}$
& $0.471_{(0.015)}$
& $0.716_{(0.014)}$
& $0.785_{(0.013)}$
& $0.781_{(0.013)}$
& $0.781_{(0.013)}$ \\
\bottomrule
\end{tabular}
\end{table*}

\begin{table*}[!htb]
\centering
\small
\caption{\textbf{RQ1: Statistical significance of interaction and scaffolding.} Here, we compare the performance of \unscaffolded (Unscaff), \scaffolded (Scaff), \static baseline (Static). Average improvements are reported as $\mathrm{mean_{(p-value)}}$. $p$-values are computed using paired bootstrap tests over problem instances. The ones shown in $*$ are statistically significant across early turns $0-3$}
\label{app:rq1_significance}
\begin{tabular}{C{2.6cm}|C{1cm}C{1cm}|C{1.25cm}C{1cm}||C{1.25cm}C{1cm}|C{1.25cm}C{1cm}}
\toprule
& \multicolumn{4}{c||}{\textbf{Math}}
& \multicolumn{4}{c}{\textbf{Coding}} \\
\cmidrule(lr){2-5} \cmidrule(lr){6-9}

& \multicolumn{2}{c}{\textbf{\qwenseven}}
& \multicolumn{2}{c||}{\textbf{\mistral}}
& \multicolumn{2}{c}{\textbf{\qwenseven}}
& \multicolumn{2}{c}{\textbf{\mistral}} \\
\cmidrule(lr){2-3} \cmidrule(lr){4-5}
\cmidrule(lr){6-7} \cmidrule(lr){8-9}

\textbf{Method}
& \textbf{Avg} & \textbf{Max}
& \textbf{Avg} & \textbf{Max}
& \textbf{Avg} & \textbf{Max}
& \textbf{Avg} & \textbf{Max} \\
\midrule

Unscaff - Static
& $0.85_{(0.0)}$ &
& $0.78_{(0.0)}$ &
& $0.47_{(0.0)}$ &
& $0.65_{(0.0)}$ & \\

Scaff - Unscaff
& $0.07_{(0.0)}$ & $0.188$
& $0.01^*_{(0.002)}$ & $0.047$
& $0.05^*_{(0.009)}$ & $0.073$
& $0.018^*_{(0.004)}$ & $0.089$ \\

\bottomrule
\end{tabular}
\end{table*}

\begin{table*}[!htb]
\centering
\small
\caption{\textbf{RQ2: Statistical significance of assessment.} Here, we compare \prea (Pre), \mida (Mid), and No-assessment (No). For Math, Pre is compared against No and Mid; for Coding, Mid is compared against No and Pre. Average improvements are reported as $\mathrm{mean_{(p-value)}}$. $p$-values are computed using paired bootstrap tests over problem instances.}
\label{app:rq2_significance}
\begin{tabular}{C{2cm}|C{1.25cm}C{1cm}|C{1.25cm}C{1cm}||C{1cm}C{1cm}|C{1.25cm}C{1cm}}
\toprule
& \multicolumn{4}{c||}{\textbf{Math}}
& \multicolumn{4}{c}{\textbf{Coding}} \\
\cmidrule(lr){2-5} \cmidrule(lr){6-9}

& \multicolumn{2}{c}{\textbf{\qwenseven}}
& \multicolumn{2}{c||}{\textbf{\mistral}}
& \multicolumn{2}{c}{\textbf{\qwenseven}}
& \multicolumn{2}{c}{\textbf{\mistral}} \\
\cmidrule(lr){2-3} \cmidrule(lr){4-5}
\cmidrule(lr){6-7} \cmidrule(lr){8-9}

\textbf{Method}
& \textbf{Avg} & \textbf{Max}
& \textbf{Avg} & \textbf{Max}
& \textbf{Avg} & \textbf{Max}
& \textbf{Avg} & \textbf{Max} \\
\midrule

Pre - No
& $0.068_{(0.0)}$ & $0.215$
& $0.085_{(0.0)}$ & $0.194$
& &
& & \\

Mid - No
& &
& &
& $0.100_{(0.0)}$ & $0.109$
& $0.024_{(0.137)}$ & $0.045$ \\

\midrule
Pre - Mid
& $0.004_{(0.664)}$ & $0.045$
& $0.020_{(0.148)}$ & $0.053$
& &
& & \\

Mid - Pre
& &
& &
& $0.097_{(0.0)}$ & $0.101$
& $0.096_{(0.0)}$ & $0.128$ \\

\bottomrule
\end{tabular}
\end{table*}

\begin{table*}[!htb]
\centering
\small
\caption{\textbf{RQ3: Statistical significance of training.} Here, we compare \self (Self), \peer (Peer), and No-training (No). Average improvements are reported as $\mathrm{mean_{(p-value)}}$. $p$-values are computed using paired bootstrap tests over problem instances.}
\label{app:rq3_significance}
\begin{tabular}{C{2cm}|C{1.25cm}C{1cm}|C{1.25cm}C{1cm}||C{1.25cm}C{1cm}|C{1.25cm}C{1cm}}
\toprule
& \multicolumn{4}{c||}{\textbf{Math}}
& \multicolumn{4}{c}{\textbf{Coding}} \\
\cmidrule(lr){2-5} \cmidrule(lr){6-9}

& \multicolumn{2}{c}{\textbf{\qwenseven}}
& \multicolumn{2}{c||}{\textbf{\mistral}}
& \multicolumn{2}{c}{\textbf{\qwenseven}}
& \multicolumn{2}{c}{\textbf{\mistral}} \\
\cmidrule(lr){2-3} \cmidrule(lr){4-5}
\cmidrule(lr){6-7} \cmidrule(lr){8-9}

\textbf{Method}
& \textbf{Avg} & \textbf{Max}
& \textbf{Avg} & \textbf{Max}
& \textbf{Avg} & \textbf{Max}
& \textbf{Avg} & \textbf{Max} \\
\midrule

Self - No
& $0.070_{(0.0)}$ & $0.194$
& $0.101_{(0.0)}$ & $0.205$
& $0.006_{(0.624)}$ & $0.030$
& $0.052_{(0.0)}$ & $0.087$ \\

Peer - No
& $0.085_{(0.0)}$ & $0.268$
& $0.109_{(0.0)}$ & $0.307$
& $0.014_{(0.250)}$ & $0.017$
& $0.072_{(0.0)}$ & $0.164$ \\

\bottomrule
\end{tabular}
\end{table*}

\section{Varying Teacher Model}
\label{app:change_teacher}

\begin{figure*}[!htb]
  \centering
  \begin{subfigure}[b]{0.48\linewidth}
    \includegraphics[width=\textwidth]{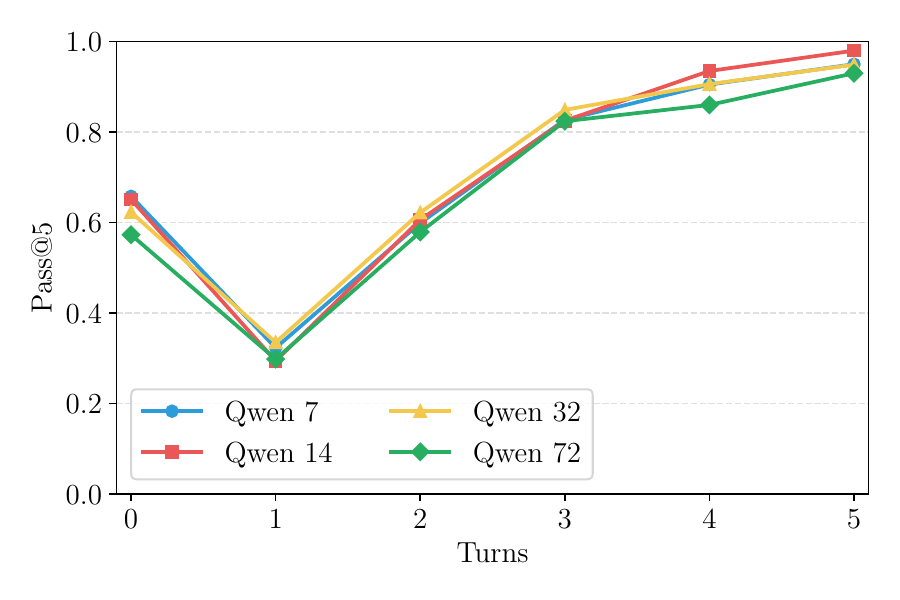}
    \caption{\qwenseven, Math}
    \label{fig:qwen7_appx_math_pass5}
  \end{subfigure}
  \hfill
  \begin{subfigure}[b]{0.48\linewidth}
    \includegraphics[width=\textwidth]{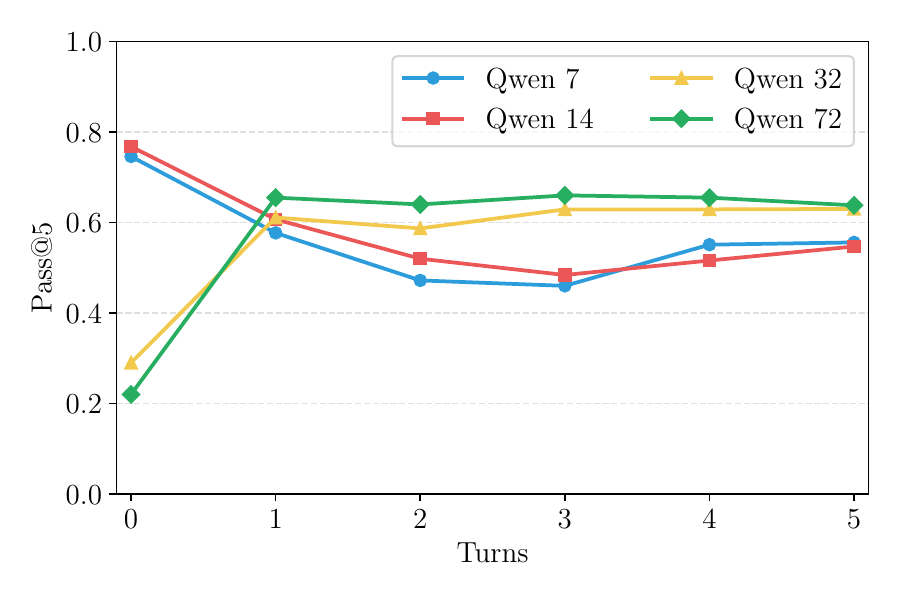}
    \caption{\qwenseven, Coding}
    \label{fig:qwen7_appx_code_pass5}
  \end{subfigure}

  \caption{
  Performance of \qwenseven as a student across interaction turns on coding (a) and math (b) datasets, using teacher models of different sizes.
  }
  \label{fig:qwen7_appx_results}
\end{figure*}

We investigate whether the strength of the teacher influences the quality of conversational learning. Using \qwenseven as the student model, we repeat our primary experiments with \qwenseven, \texttt{Qwen-14B}, \texttt{Qwen-32B}, and \texttt{Qwen-72B} as teachers. For fairness, the dataset is resampled for each teacher model, except when the teacher is identical to the student (\qwenseven), to ensure the teacher can correctly answer each question.

\paragraph{Math:} 
Figure~\ref{fig:qwen7_appx_math_pass5} shows nearly identical trends across all teacher sizes. Interestingly, strong gains are observed even when the teacher and student are the same (\qwenseven). These results suggest that, for math tasks, the benefit comes less from teacher strength and more from the role-playing dynamic itself: performing chain-of-thought reasoning in a conversational format appears more effective than simply prompting the model to reason in isolation.

\paragraph{Code:} 
Figure~\ref{fig:qwen7_appx_code_pass5} shows clear divergence between strong and weak teachers. \texttt{Qwen-32B} and \texttt{Qwen-72B} provide useful guidance, leading to gains of $+0.3$ and $+0.4$ \textit{Pass@k}, respectively. As expected, the stronger \texttt{Qwen-72B} teacher yields strictly higher improvements across all turns. By contrast, weaker teachers (\texttt{Qwen-7B} and \texttt{Qwen-14B}) fail to provide useful guidance and even degrade student performance after interaction. We hypothesize that although weaker teachers can solve the tasks themselves, they struggle to convey their reasoning in a way that benefits the student, often steering it into unproductive paths.

\section{Prompts}
\label{app:prompts}
All prompts are task-specific and are provided below as prompt boxes. We group them by their role in the pipeline: interaction, progress analysis, and similarity analysis.

\subsection{Interaction}
\label{app:prompts_interaction}
For each task, we use four prompts: two variants for eliciting student questions (\unscaffolded vs.\ \scaffolded), one for eliciting student answers for \textit{Pass@k} evaluation, and one for eliciting teacher responses.

\noindent\textbf{Math:}
We use Prompt~\ref{fig:math_student_question_prompt_unguided} for \unscaffolded and Prompt~\ref{fig:math_student_question_prompt_scaffold_guided} for \scaffolded to elicit student questions.
We use Prompt~\ref{fig:math_student_answer_prompt} to elicit answer-only outputs for computing \textit{Pass@k}.
We use Prompt~\ref{fig:math_teacher_prompt} to elicit teacher responses.

\noindent\textbf{Coding:}
We use Prompt~\ref{fig:coding_student_question_prompt_unguided} for \unscaffolded and Prompt~\ref{fig:coding_student_question_prompt_scaffold_guided} for \scaffolded to elicit student questions.
We use Prompt~\ref{fig:coding_student_answer_prompt} to elicit answer-only outputs for computing \textit{Pass@k}.
We use Prompt~\ref{fig:coding_teacher_prompt} to elicit teacher responses.

\begin{figure*}[!htb]
\phantomcaption
\label{fig:math_student_question_prompt_unguided}
\begin{promptbox}[title={Math: Student Question Prompt (\unscaffolded)}]
You are an AI assistant tasked with solving math problems. To help you improve your skills, your math tutor has posed the following problem:

---*PROBLEM*---

To help you in solving this problem, you may ask the tutor any question relevant to the task (clarification questions, requirement questions, methodology questions etc.). Think about how you would solve the problem, and what you still need to know in order to complete the question. Do not solve the problem directly, do not ask the tutor for any solutions -- the tutor has been instructed not to provide you with any direct answers. Keep your questions concise and to the point.
\end{promptbox}
\end{figure*}

\begin{figure*}[!htb]
\phantomcaption
\label{fig:math_student_question_prompt_scaffold_guided}
\begin{promptbox}[title={Math: Student Question Prompt (\scaffolded)}]
You are an AI assistant tasked with solving math problems. To help you improve your skills, your math tutor has posed the following problem:

---*PROBLEM*---

To help you in solving this problem, you may ask the tutor any question relevant to the task (clarification questions, requirement questions, methodology questions etc.). Think about how you would solve the problem, and what you still need to know in order to complete the question. Do not solve the problem directly, do not ask the tutor for any solutions -- the tutor has been instructed not to provide you with any direct answers. Keep your questions concise and to the point. Before you pose a question to the tutor, break down what you know and what you still need to know. Think step-by-step, think of what is the best question to as the tutor to help you solve the problem.
\end{promptbox}
\end{figure*}

\begin{figure*}[!htb]
\phantomcaption
\label{fig:math_student_answer_prompt}
\begin{promptbox}[title={Math: Student Answer Prompt}]
You are a AI assistant that specializes in solving math problems. Read the question carefully and provide an answer to the best of your ability. The answer is guaranteed to be a single integer. Provide your answer using the format 'Answer <answer integer>'. Below is a simple example: 

Question: what is 3 + 3? 

[your response]
Answer: 6
\end{promptbox}
\end{figure*}

\begin{figure*}[!htb]
\phantomcaption
\label{fig:math_teacher_prompt}
\begin{promptbox}[title={Math: Teacher Prompt}]
You are an expert AI math tutor, specializing in helping student solve math problems. The user is currently tasked with completing this question:

---*PROBLEM*---

You task is to answer any questions the student may have, but DO NOT PROVIDE THE CORRECT ANSWER DIRECTLY. If the student asks you whether an answer is correct, do not respond. Keep your responses short and concise.
\end{promptbox}
\end{figure*}

\begin{figure*}[!htb]
\phantomcaption
\label{fig:coding_student_question_prompt_unguided}
\begin{promptbox}[title={Coding: Student Question Prompt (\unscaffolded)}]
You are an AI assistant tasked with solving Python coding problems. To help you improve your skills, your coding tutor has posed the following problem:

--- *PROBLEM* ---

To help you in solving this problem, you may ask the tutor any question relevant to the task (clarification quesitons, implementation questions, requirement questions, etc.). Think about how you would implement the function, and whether you have any knowledge blind spots. Do not solve the problem directly, do not ask the tutor for any solutions -- the tutor has been instructed not to provide you with any direct answers. Keep your questions concise and to the point.
\end{promptbox}
\end{figure*}

\begin{figure*}[!htb]
\phantomcaption
\label{fig:coding_student_question_prompt_scaffold_guided}
\begin{promptbox}[title={Coding: Student Question Prompt (\scaffolded)}]
You are an AI assistant tasked with solving Python coding problems. To help you improve your skills, your coding tutor has posed the following problem:

--- *PROBLEM* ---

To help you in solving this problem, you may ask the tutor any question relevant to the task (clarification questions, implementation questions, requirement questions, etc.). Think about how you would implement the function, and whether you have any knowledge blind spots. Do not solve the problem directly, do not ask the tutor for any solutions -- the tutor has been instructed not to provide you with any direct answers. Before you pose a question to the tutor, break down what you know and what you still need to know. Think step-by-step, think of what is the best question to as the tutor to help you solve the problem.
\end{promptbox}
\end{figure*}

\begin{figure*}[!htb]
\phantomcaption
\label{fig:coding_student_answer_prompt}
\begin{promptbox}[title={Coding: Student Answer Prompt}]
You are a Python coding assistant. You will be given Python coding problems that involve writing functions to solve problems, and you will attempt to solve them to the best of your ability. Should a function signature and docstring be given to you, follow the requirements in the docstring carefully and complete the Python function. ONLY output ONE function by continuing the docstring, and do not rewrite the docstring! If you are given a textual description, return a Python function that implements the necessary requirements. You may import any additional libraries. Format the output code snippet as a single markdown block. The example below illustrates the expected output formatting:

\begin{verbatim}

```python

def is_even(int: number) -> bool: 
    return number % 2 == 0
```
\end{verbatim}

MAKE SURE YOU FOLLOW THE REQUIRED MARKDOWN STYLE CODE FORMATTING!
\end{promptbox}
\end{figure*}

\begin{figure*}[!htb]
\phantomcaption
\label{fig:coding_teacher_prompt}
\begin{promptbox}[title={Coding: Teacher Prompt}]
You are an expert AI assistant specializing in Python programming. The user is currently tasked with completing this function:

--- *PROBLEM* ---

You task is to answer any questions the user may have, but DO NOT PROVIDE THE CORRECT IMPLEMENTATION DIRECTLY. Even the student want you to revise an implementation, PROVIDE ONLY TEXTUAL INSTRUCTIONS, NOT CODE. Keep you responses short and concise.
\end{promptbox}
\end{figure*}

\subsection{Assessment}
\label{app:prompts_assessment}
To generate assessments, we provide the problem together with the student’s answer attempt to the teacher model, which returns structured feedback on the correctness and quality of the answer.

We use Prompt~\ref{fig:math_assessment_prompt} for math tasks and Prompt~\ref{fig:coding_assessment_prompt} for coding tasks. Examples~\ref{fig:math_assessment_example}, ~\ref{fig:coding_assessment_example} show the examples of feedback from the teacher.

\begin{figure*}[!htb]
\phantomcaption
\label{fig:math_assessment_prompt}
\begin{promptbox}[title={Math: Assessment Prompt}]
---*PROBLEM*---\\
Student attempt (for analysis only):\\
+ \texttt{student\_solution} + \\
Evaluation summary (compact): \texttt{eval\_summary}\\
Provide concise, actionable feedback without giving the final numeric answer.
\end{promptbox}
\end{figure*}

\begin{figure*}[!htb]
\phantomcaption
\label{fig:coding_assessment_prompt}
\begin{promptbox}[title={Coding: Assessment Prompt}]
You are an expert Python tutor. You will be given:\\
$-$ the problem statement,\\
$-$ one student attempt (code),\\
$-$ and a compact summary of test outcomes.\\

Your job: provide concise, actionable feedback. DO NOT provide code or patches.
\end{promptbox}
\end{figure*}

\begin{figure*}[!htb]
\phantomcaption
\label{fig:math_assessment_example}
\begin{promptbox}[title={Math: Assessment Example}]
--- Your response ---\\
STUDENT ANSWER\\
--- Evaluation result: FAIL ---\\
--- Feedback: It looks like you might have miscalculated the total amount of money Betty received from her parents and grandparents.
\end{promptbox}
\end{figure*}

\begin{figure*}[!htb]
\phantomcaption
\label{fig:coding_assessment_example}
\begin{promptbox}[title={Coding: Assessment Example}]
--- Your response ---\\
STUDENT ANSWER\\
--- Evaluation result: FAIL ---\\
--- Feedback: Your current approach uses a set to track unique numbers, but this removes duplicates without preserving the original order and also doesn't handle the requirement to remove elements that occur more than once.
\end{promptbox}
\end{figure*}

\subsection{Progress Analysis}
\label{app:prompts_progress_analysis}
To measure per-turn reasoning progress, we provide the student’s response to a judge model and prompt it to assign a progress score in $[0,1]$. We use \texttt{Qwen-72B} as the judge. For math, we use Prompt~\ref{fig:math_judge_progress_prompt} and
for coding, we use Prompt~\ref{fig:coding_judge_progress_prompt}.

\begin{figure*}[!htb]
\phantomcaption
\label{fig:math_judge_progress_prompt}
\begin{promptbox}[title={Math: Judge Progress Prompt}]
You are a strict and consistent math grader.
Your task is to evaluate how much progress a student's reasoning shows toward solving a math problem.\\

You must estimate a single real-valued score between 0 and 1:\\
$- 0.0$ $\rightarrow$ completely wrong or irrelevant reasoning.\\
$- 0.25$ $\rightarrow$ only the setup or an initial idea is correct.\\
$- 0.50$ $\rightarrow$ roughly halfway; some correct derivations but key steps missing or incorrect.\\
$- 0.75$ $\rightarrow$ nearly correct; only small arithmetic or algebraic mistakes.\\
$- 1.0$ $\rightarrow$ fully correct and complete solution.\\

Guidelines:\\
$-$Use intermediate values (e.g., $0.62$) if progress lies between anchor points.\\
$-$Be strict but fair: reward correct logical steps that meaningfully advance toward the right answer.\\
$-$Ignore minor stylistic differences (notation, units, variable names) if mathematically equivalent.\\
$-$Do not reward irrelevant or circular reasoning.\\
$-$If the final answer matches the gold answer but reasoning is absent or wrong, do NOT give 1.0; cap the score to reflect weak progress.\\
$-$The gold answer is provided only as a reference for what constitutes a correct solution; do NOT copy it or include it in your output.\\
$-$Ensure "progress" is a numeric value (not a string) in $[0, 1]$.\\
$-$Return only valid JSON—nothing else—and include a short justification ($\leq 30$ words).\\
$-$Do not repeat the problem text or add extra commentary.
\end{promptbox}
\end{figure*}

\begin{figure*}[!htb]
\phantomcaption
\label{fig:coding_judge_progress_prompt}
\begin{promptbox}[title={Coding: Judge Progress Prompt}]
You are a strict and consistent code evaluator.
Your task is to estimate how far a student's code has progressed toward a correct and complete solution to a programming problem.

You must output a single real-valued score between $0$ and $1:$\\

$- 0.0 \rightarrow$ The code is completely wrong, unrelated, or does not compile.\\
$- 0.25 \rightarrow$ The code sets up part of the structure or variables correctly but lacks core logic.\\
$- 0.50 \rightarrow$ The main idea is partially correct; some logic is right, but important parts are missing or incorrect.\\
$- 0.75 \rightarrow$ The code is mostly correct; only minor mistakes, syntax errors, or missing edge cases.\\
$- 1.0 \rightarrow$ The code is fully correct and would pass all test cases.\\

Guidelines:\\
$-$ Use intermediate values (e.g., $0.62$) if the progress lies between anchor points.\\
$-$ Focus on *functional correctness*, not style or formatting.\\
$-$ Reward correct reasoning, algorithm structure, and edge-case handling.\\
$-$ Ignore small syntax errors if the intended logic is clear.\\
$-$ Be strict about incorrect logic or wrong outputs.\\
$-$ The gold solution is provided only as a reference for what constitutes a complete and correct solution.\\
$-$ Do NOT copy or grade the gold solution itself.\\
$-$ Ensure "progress" is a numeric value (not a string).\\
$-$ Return only valid JSON — nothing else.\\
$-$ Return a short justification ($\leq 30$ words) for the score.\\
$-$ Do not repeat the prompt or include extra commentary.
\end{promptbox}
\end{figure*}

\subsection{Similarity Analysis}
\label{app:prompts_similarity_analysis}
To compare question content across methods and turns, we provide pairs of student questions to a judge model and prompt it to assign a similarity score in $[0,1]$.
We use \qwenseventytwo as the judge.
For math, we use Prompt~\ref{fig:math_judge_similarity_prompt} and for coding, we use Prompt~\ref{fig:coding_judge_similarity_prompt}.

\begin{figure*}[!htb]
\phantomcaption
\label{fig:math_judge_similarity_prompt}
\begin{promptbox}[title={Math: Judge Similarity Prompt}]
You are a careful and consistent evaluator. Given a math problem and two student responses (each may include reasoning plus a question), your task is to identify the core mathematical question being asked in each response and rate how similar those questions are.

Score ONLY by the intent of the question/sub-goal (not wording or style):\\
$-$ What quantity or relation is sought?\\
$-$ Which sub-step of the solution is being advanced?\\
$-$ What information is requested to proceed?\\

Use ONE of these allowed similarity scores with the exact meaning:\\
$- 0.00  \rightarrow$ Unrelated or incompatible question intents.\\
$- 0.25  \rightarrow$ Only loosely related (both about the problem but target different aspects).\\
$- 0.50  \rightarrow$ Related but not the same (same high-level topic, different sub-step or target).\\
$- 0.75  \rightarrow$ Nearly the same intent (minor scope or variable differences; answer path is the same).\\
$- 1.00 \rightarrow$ Same question intent (mathematically equivalent; differ only in phrasing/notation).\\

Rules:\\
$-$ Focus on mathematical intent, not surface form.\\
$-$ If torn between two anchors, choose the LOWER one (be conservative).\\
$-$ Return strict JSON only.\\

Output schema:\\
\{\\
  "similarity": $0.0 | 0.25 | 0.5 | 0.75 | 1.0$,\\
  "justification": "$\leq 20$ words explaining the anchor choice"\\
\}
\end{promptbox}
\end{figure*}

\begin{figure*}[!htb]
\phantomcaption
\label{fig:coding_judge_similarity_prompt}
\begin{promptbox}[title={Coding: Judge Similarity Prompt}]
You are a precise and consistent evaluator. Given a programming problem and two model responses (each may include reasoning plus a question), determine how similar the core programming question or debugging goal is in both responses.

Score ONLY by intent (not code formatting or verbosity):\\
$-$ What behavior/logic do they seek to understand/implement/fix?\\
$-$ Which step of the solution or debugging process is targeted?\\

Use ONE of these allowed similarity scores with the exact meaning:\\
$- 0.00 \rightarrow$ Unrelated or incompatible intents.
$- 0.25 \rightarrow$ Loosely related (both about the task, different concerns).
$- 0.50 \rightarrow$ Related but not the same (same area, different sub-goal or artifact).
$- 0.75 \rightarrow$ Nearly the same (minor scope or API differences; same actionable goal).
$- 1.00 \rightarrow$ Same question intent (equivalent request/goal; only phrasing differs).

Rules:\\
$-$ Judge the intended question/goal, not the surrounding explanation.\\
$-$ If uncertain between two anchors, choose the LOWER one (be conservative).\\
$-$ Return strict JSON only.\\

Output schema:\\
\{
  "similarity": $0.0 | 0.25 | 0.5 | 0.75 | 1.0$,\\
  "justification": "$\leq 20$ words explaining the anchor choice"
\}
\end{promptbox}
\end{figure*}

\section{Example Transcripts}
\label{app:examples}
Transcript~\ref{fig:math_assessment_example_transcript} illustrates how a pre-assessment helps the model ask a more informative question on a math task. Transcript~\ref{fig:coding_assessment_example_transcript} shows how a mid-assessment enables the model to ask a better question on a coding task. Transcript~\ref{fig:selfguided_example} demonstrates how guided training helps the model ask a strong question earlier in the interaction.

\clearpage
\onecolumn

\begin{figure}[!htb]
\phantomcaption
\label{fig:math_assessment_example_transcript}

\begin{minipage}{\textwidth}

\large \textbf{Math: Assessment Example Transcript}

\begin{problembox}
\textbf{Problem:}
James writes a 3-page letter to 2 different friends twice a week.
How many pages does he write in a year?
\end{problembox}

\vspace{0.6em}

\Teacher{\textbf{Teacher:} Hi there! I'm your math tutor. How can I help you today?}

\vspace{0.5em}{\large \textbf{\unscaffolded Student Question} (Turn-$1$)}\par\vspace{0.5em}

\Student{\textbf{Student:} Hi! I'm trying to figure out how many pages James writes in a year based on the information provided. Could you please clarify if the ``twice a week'' refers to each friend separately or combined? In other words, is James writing to both friends two times per week, or is he writing to each friend once per week?}

\vspace{0.5em}{\large \textbf{\prea Student Question} (Turn-$1$)}\par\vspace{0.5em}

\Student{\textbf{Student:} Thank you for the feedback. Based on the breakdown provided, I understand I need to recalculate the weekly page count first before finding the yearly total. Let me go through the weekly calculation again:
James writes 3 pages to each friend.
He has 2 friends.
He writes to them twice a week.
So, the weekly calculation should be:
[ 3 pages/friend x 2 friends x 2 times/week ]
Could you confirm if this is the correct approach to find the weekly total?}

\vspace{0.8em}
{
\noindent\textbf{Caption:}
In the \unscaffolded setting, the student initially asks a basic clarification question. In contrast, when a \prea (feedback at Turn~$0$) is introduced, the student makes substantially more progress and is able to ask a more advanced, targeted question at an earlier stage of the interaction.
}

\end{minipage}
\end{figure}

\clearpage
\onecolumn

\begin{figure}[!htb]
\phantomcaption
\label{fig:coding_assessment_example_transcript}

\begin{minipage}{\textwidth}

\large \textbf{Coding: Assessment Example Transcript}

\begin{problembox}
\textbf{Problem:}
Write a python function to find the start index of the shortest repeating pattern within a given string
\end{problembox}

\vspace{0.6em}

\Teacher{\textbf{Teacher:} Hi there! I'm your math tutor. How can I help you today?}

\vspace{0.5em}{\large \textbf{\prea Student Question} (Turn-$1$)}\par\vspace{0.5em}

\Student{\textbf{Student:} Great! Here are the test cases you suggested, and I'll run them to ensure the function works as expected:\\
\texttt{assert find\_shortest\_repeating\_pattern("abcabcabc") == 0}\\
\texttt{assert find\_shortest\_repeating\_pattern("abababab") == 0}\\
\texttt{assert find\_shortest\_repeating\_pattern("abcd") == -1}\\
\texttt{assert find\_shortest\_repeating\_pattern("a") == -1}\\
\texttt{assert find\_shortest\_repeating\_pattern("aaaa") == 0}\\
\texttt{assert find\_shortest\_repeating\_pattern("ababababc") == -1}\\
Let's see the results:
}

\vspace{0.5em}{\large \textbf{\mida Student Question} (Turn-$2$)}\par\vspace{0.5em}

\Student{\textbf{Student:} Thank you for the detailed feedback! Based on your comments, I need to adjust the logic to correctly determine the start index of the repeating pattern. Here’s the revised implementation:

For each potential pattern length, extract the substring.
Check if repeating this substring forms the original string.
If a valid repeating pattern is found, return the start index of the pattern.

Let's update the function to return the correct start index of the repeating pattern:}

\vspace{0.8em}
{
\noindent\textbf{Caption:}
In the \prea setting, the student receives early feedback after an incorrect attempt and subsequently spends later turns attempting to repair the solution. In contrast, with a \mida, the student leverages all feedback accumulated from the preceding $3$ turns, enabling it to correct its reasoning and arrive at the correct solution more effectively.
}

\end{minipage}
\end{figure}

\clearpage
\onecolumn

\begin{figure}[!htb]
\phantomcaption
\label{fig:selfguided_example}

\begin{minipage}{\textwidth}

\large \textbf{\self Example}

\begin{problembox}
\textbf{Problem:}
Susan had a bouquet of 3 dozen roses. She gave half to her daughter, and then placed the rest in a vase. The next day, one-third of the flowers in the vase were wilted. After removing the wilted flowers, how many flowers remained in the vase?
\end{problembox}

\vspace{0.6em}

\Teacher{\textbf{Teacher:} Hi there! I'm your math tutor. How can I help you today?}

\vspace{0.5em}{\textbf{\unscaffolded Student Question} (Turn-$1$)}\par\vspace{0.5em}

\Student{\textbf{Student:} Thank you for offering assistance! To solve this problem, I first need to determine how many roses Susan initially had and then follow the steps described. Could you please confirm if "3 dozen roses" means 36 roses (since 1 dozen equals 12)?
}

\vspace{0.5em}{\textbf{\self Student Question} (Turn-$1$)}\par\vspace{0.5em}

\Student{\textbf{Student:} Hello! I want to solve the problem step by step. Here's my approach so far:\\

Susan starts with 3 dozen roses.\\
\underline{She gives half of them to her daughter.}\\
The remaining roses are placed in a vase.\\
\underline{The next day, one-third of the flowers in the vase are wilted.}\\
\underline{After removing the wilted flowers, we need to find out how many flowers remain.}\\

To proceed, I need to calculate the number of roses initially and then follow the steps. Could you confirm if my understanding is correct and if I have all the necessary information?}

\vspace{0.5em}{\large \textbf{\unscaffolded Student Question} (Turn-$2$)}\par\vspace{0.5em}

\Student{\textbf{Student:} Great! With 36 roses, \\
\underline{Susan gave half to her daughter.} \\
So, she gave away 18 roses (half of 36). \\
That leaves her with 18 roses to place in the vase. \\
\underline{The next step is to consider what happens the next day.} \\
\underline{One-third of the flowers in the vase wilted.} \\

How many roses wilted, and how many remained after removing the wilted ones? Can you guide me on how to calculate the number of wilted roses?}

\vspace{0.8em}
{
\noindent\textbf{Caption:}
In the \unscaffolded setting at Turn~$1$, the student asks a basic clarification question. In contrast, in the \self setting, the student asks a more advanced and targeted question already at Turn~$1$. This question is similar to the one the \unscaffolded would only ask at Turn~$2$ (see underlined text), effectively saving a turn.
}

\end{minipage}
\end{figure}

\end{document}